\documentclass[10pt,journal,compsoc]{IEEEtran}

\ifCLASSOPTIONcompsoc 
	\usepackage[nocompress]{cite} 
\else
	\usepackage{cite} 
\fi

\ifCLASSOPTIONcaptionsoff
	\usepackage[nomarkers]{endfloat}
	\let\MYoriglatexcaption\caption
	\renewcommand{\caption}[2][\relax]{\MYoriglatexcaption[#2]{#2}}
\fi

\ifCLASSINFOpdf 
	\usepackage[pdftex]{graphicx}
	\DeclareGraphicsExtensions{.pdf}
\else
\fi

\usepackage{amsmath} 
\usepackage{algorithmic} 

\usepackage{array} 
\usepackage{multirow}
\usepackage{multicol}
\usepackage{threeparttable}
\usepackage{booktabs}
\usepackage{enumitem} 

\usepackage{subfigure}
\usepackage{color}
\usepackage{url}
\usepackage{amssymb}

\begin{document}

\title{MFQE 2.0: A New Approach for Multi-frame Quality Enhancement on Compressed Video}

	\author{
		Qunliang~Xing*,
		Zhenyu~Guan*,
    	Mai~Xu,~\IEEEmembership{Senior~Member,~IEEE,}
    	Ren~Yang,
    	Tie~Liu
    	and~Zulin~Wang%
        \IEEEcompsocitemizethanks{
            \IEEEcompsocthanksitem Accepted by IEEE Transactions on Pattern Analysis and Machine Intelligence. DOI: 10.1109/TPAMI.2019.2944806.
            \IEEEcompsocthanksitem Q. Xing and Z. Guan contribute equally to this paper.
            \IEEEcompsocthanksitem Corresponding author: Mai Xu.
            }
        }



\IEEEtitleabstractindextext{%
\begin{abstract}
The past few years have witnessed great success in applying deep learning to enhance the quality of compressed image/video.
The existing approaches mainly focus on enhancing the quality of a single frame, not considering the similarity between consecutive frames.
Since heavy fluctuation exists across compressed video frames as investigated in this paper, frame similarity can be utilized for quality enhancement of low-quality frames given their neighboring high-quality frames.
This task is Multi-Frame Quality Enhancement (MFQE).
Accordingly, this paper proposes an MFQE approach for compressed video, as the first attempt in this direction.
In our approach, we firstly develop a Bidirectional Long Short-Term Memory (BiLSTM) based detector to locate Peak Quality Frames (PQFs) in compressed video.
Then, a novel Multi-Frame Convolutional Neural Network (MF-CNN) is designed to enhance the quality of compressed video, in which the non-PQF and its nearest two PQFs are the input. In MF-CNN, motion between the non-PQF and PQFs is compensated by a motion compensation subnet.
Subsequently, a quality enhancement subnet fuses the non-PQF and compensated PQFs, and then reduces the compression artifacts of the non-PQF.
Also, PQF quality is enhanced in the same way.
Finally, experiments validate the effectiveness and generalization ability of our MFQE approach in advancing the state-of-the-art quality enhancement of compressed video.
The code is available at \textit{https://github.com/RyanXingQL/MFQEv2.0.git}.
\end{abstract}

}

\maketitle

\IEEEdisplaynontitleabstractindextext

\IEEEpeerreviewmaketitle

\IEEEraisesectionheading{\section{Introduction}\label{sec:introduction}}

\IEEEPARstart{D}{uring} the past decades, there has been a considerable increase in the popularity of video over the Internet. According to Cisco Data Traffic Forecast \cite{Cisco}, video generates $60\%$ of Internet traffic in 2016, and this figure is predicted to reach $78\%$ by 2020. When transmitting video over the bandwidth-limited Internet, video compression has to be applied to significantly save the coding bit-rate. However, the compressed video inevitably suffers from compression artifacts, which severely degrade the Quality of Experience (QoE) \cite{seshadrinathan2010study,li2015weight,tan2016video,bampis2017study,yang2018saliency}. Besides, such artifacts may reduce the accuracy for tasks of classification and recognition. It is verified in \cite{gupta2005restoration,hennings2008simultaneous,nishiyama2009facial,zhang2011close} that compression quality enhancement can improve the performance of classification and recognition. Therefore, there is a pressing need to study on quality enhancement for compressed video.

\begin{figure}[!t]
	\vspace{+1em}
	\centering
	\includegraphics[width=1\linewidth]{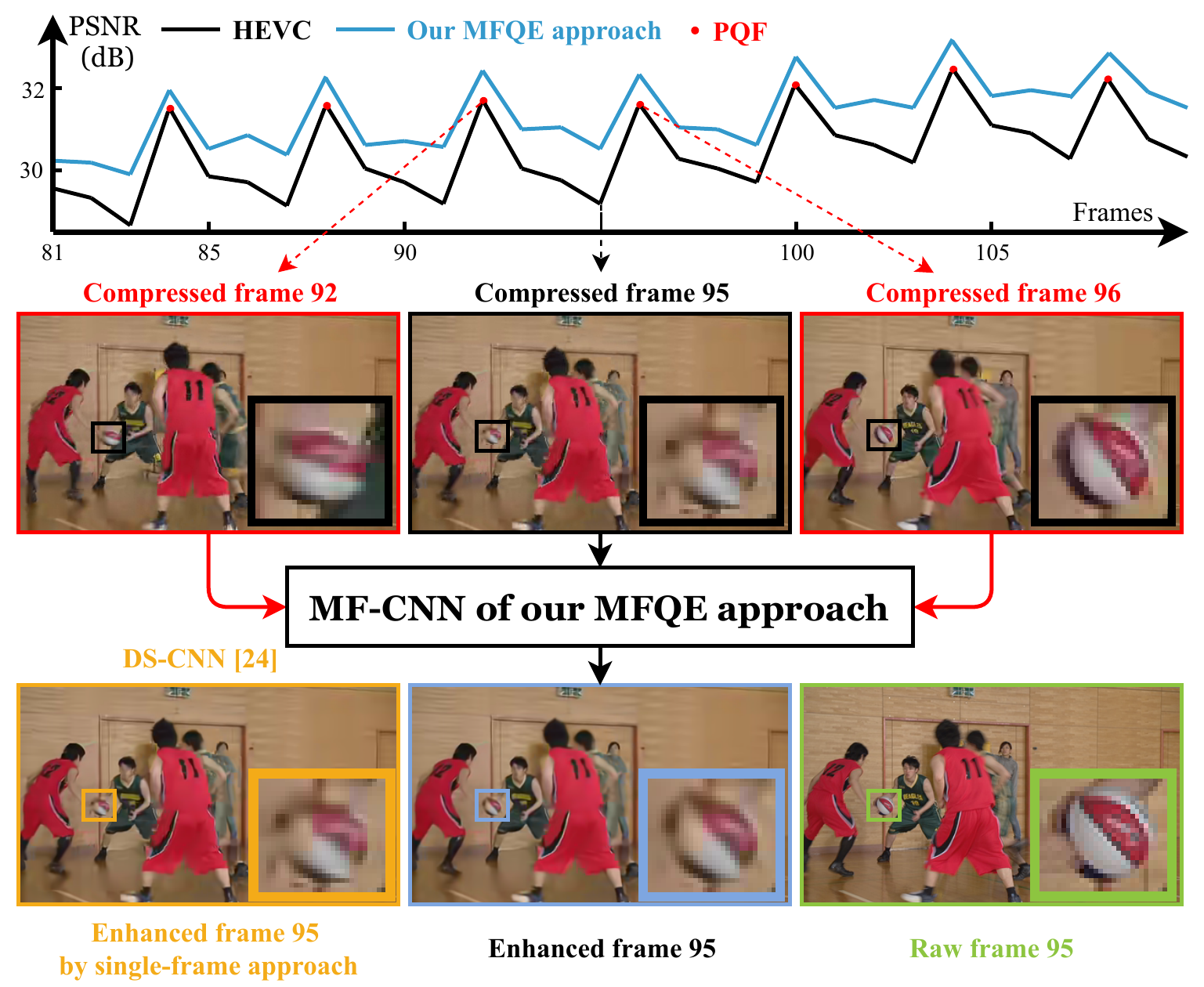}
	\caption{\footnotesize{An example for quality fluctuation (top) and quality enhancement performance (bottom).}}
	\label{fig1}
	\vspace{+0.0em}
\end{figure}

Recently, extensive works were conducted for enhancing the visual quality of compressed image and video \cite{liew2004blocking,foi2007pointwise,wang2013adaptive,jancsary2012loss,jung2012image,chang2014reducing,dong2015compression,Guo2016Building, wang2016d3,Zhang2017Beyond,li2017efficient,Cavigelli2017CAS,tai2017memnet, yang2017decoder, yang2018enhancing}.
For example, Dong \textit{et al.} \cite{dong2015compression} designed a four-layer Convolutional Neural Network (CNN) \cite{lecun1998gradient}, named AR-CNN, which considerably improves the quality of JPEG images.
Then, Denoising CNN (DnCNN) \cite{Zhang2017Beyond}, which applies residual learning strategy, was proposed for image denoising, image super-resolution and JPEG quality enhancement. Later, Yang \textit{et al.} \cite{yang2017decoder, yang2018enhancing} designed a Decoder-side Scalable CNN (DS-CNN) for video quality enhancement. The DS-CNN structure is composed of two subnets, aiming at reducing intra- and inter-coding distortion, respectively. However, when processing a single frame, all existing quality enhancement approaches do not take any advantage of the information provided by neighboring frames, and thus their performance is severely limited. As Fig.\thinspace\ref{fig1} shows, the quality of compressed video dramatically fluctuates across frames. Therefore, it is possible to use the high-quality frames (i.e., Peak Quality Frames, called PQFs\footnote{PQF is defined as the frame whose quality is higher than both its previous frame and subsequent frame.}) to enhance the quality of their neighboring low-quality frames (non-PQFs). This can be seen as Multi-Frame Quality Enhancement (MFQE), similar to multi-frame super-resolution \cite{Kappeler2016Video,Li2017Video,Caballero_2017_CVPR}.

This paper proposes an MFQE approach for compressed video. Specifically, we investigate that there exists large quality fluctuation in consecutive frames, for video sequences compressed by almost all compression standards.
Thus, it is possible to improve the quality of a non-PQF with the help of its neighboring PQFs.
To this end, we first train a Bidirectional Long Short-Term Memory (BiLSTM) based model as a no-reference method to detect PQFs.
Then, a novel Multi-Frame CNN (MF-CNN) architecture is proposed for non-PQF quality enhancement, which takes both the current non-PQF and its adjacent PQFs as input.
Our MF-CNN includes two components, i.e., Motion Compensation subnet (MC-subnet) and Quality Enhancement subnet (QE-subnet).
The MC-subnet is developed to compensate motion between current non-PQF and its adjacent PQFs.
The QE-subnet, with a spatio-temporal architecture, is designed to extract and merge the features of current non-PQF and compensated PQFs.
Finally, the quality of the current non-PQF can be enhanced by QE-subnet which takes advantage of higher quality information provided by its adjacent PQFs.
For example, as shown in Fig.\thinspace\ref{fig1}, the current non-PQF (frame 95) and its nearest two PQFs (frames 92 and 96) are both fed into MF-CNN in our MFQE approach.
As a result, the low-quality content (basketball) in non-PQF (frame 95) can be enhanced upon essentially the same but qualitatively better content in neighboring PQFs (frames 92 and 96).
Moreover, Fig.\thinspace\ref{fig1} shows that our MFQE approach also mitigates the quality fluctuation, due to the considerable quality improvement of non-PQFs.
Note that our MFQE approach is also used for reducing compression artifacts of PQFs by using neighboring PQFs to enhance the quality of the currently processed PQF.

This work is an extended version of our conference paper \cite{yang2018multi} (called MFQE 1.0 in this paper) with additional works and substantial improvements, thus called MFQE 2.0 (called MFQE in this paper for simplicity).
The extension is as follows.
(1) We enlarge our database in \cite{yang2018multi} from 70 to 160 uncompressed videos. On this basis, more thorough analyses of the compressed video are conducted.
(2) We develop a new PQF-detector, which is based on BiLSTM instead of the support vector machine (SVM) in \cite{yang2018multi}. Our new detector is capable of extracting both spatial and temporal information of PQFs, leading to a boost in $F_1$-score of PQF detection from $91.1\%$ to $98.2\%$.
(3) We advance our QE-subnet by introducing the multi-scale strategy, batch normalization \cite{Ioffe2015Batch} and dense connection \cite{huang2017densely}, rather than the conventional design of CNN in \cite{yang2018multi}.
Besides, we develop a lightweight structure for the QE-subnet to accelerate the speed of video quality enhancement.
Experiments show that the average Peak Signal-to-Noise Ratio (PSNR) improvement on 18 sequences selected by \cite{ohm2012comparison}  largely increases from 0.455 dB to 0.562 dB (i.e., $23.5\%$ improvement), while the number of parameters substantially reduces from 1,787,547 to 255,422 (i.e., $85.7\%$ saving), resulting in at least 2 times acceleration of quality enhancement.
(4) More extensive experiments are provided to validate the performance and generalization ability of our MFQE approach.

\begin{figure*}[!t]
	\vspace{-0.0em}
	\centering
	\includegraphics[width = 6.8in]{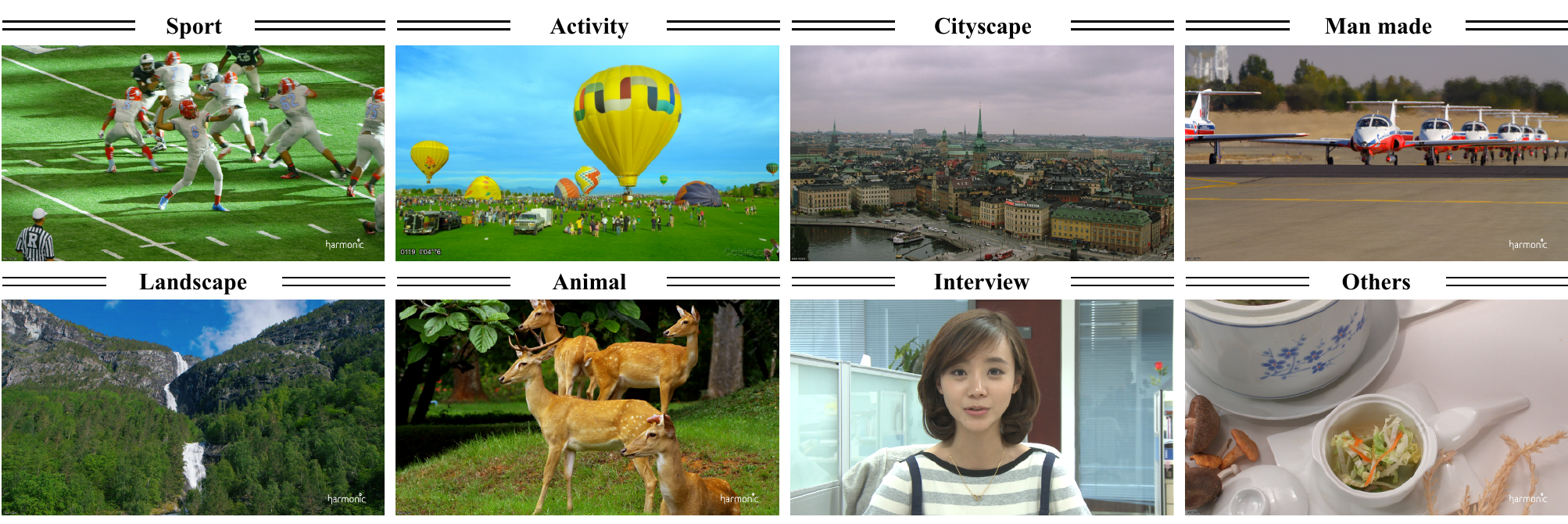}
	\caption{Examples of video sequences in our enlarged database.}
	\label{diversity}
	\vspace{-0.0em}
\end{figure*}

\begin{figure*}[h]
	\vspace{-0.0em}
	\centering
	\begin{minipage}[t]{2.3in}
		\centering
		\includegraphics[width = 2.3in]{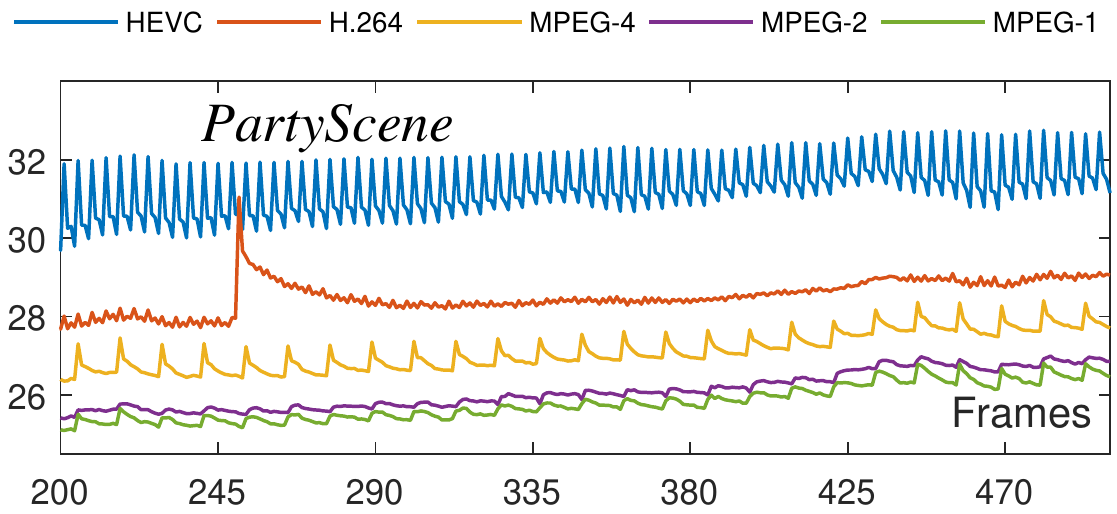}
	\end{minipage}
	\begin{minipage}[t]{2.3in}
		\centering
		\includegraphics[width = 2.3in]{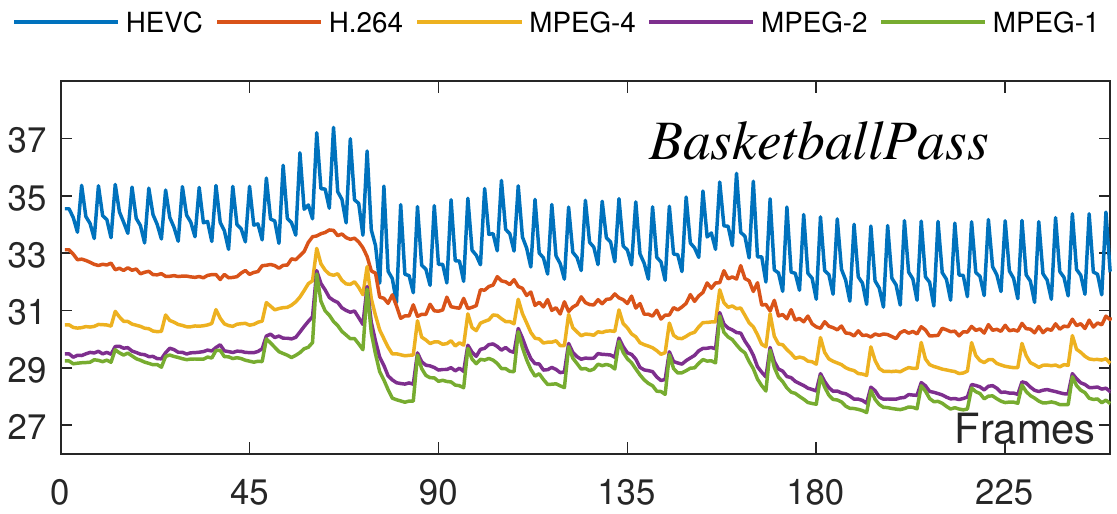}
	\end{minipage}
	\begin{minipage}[t]{2.3in}
		\centering
		\includegraphics[width = 2.3in]{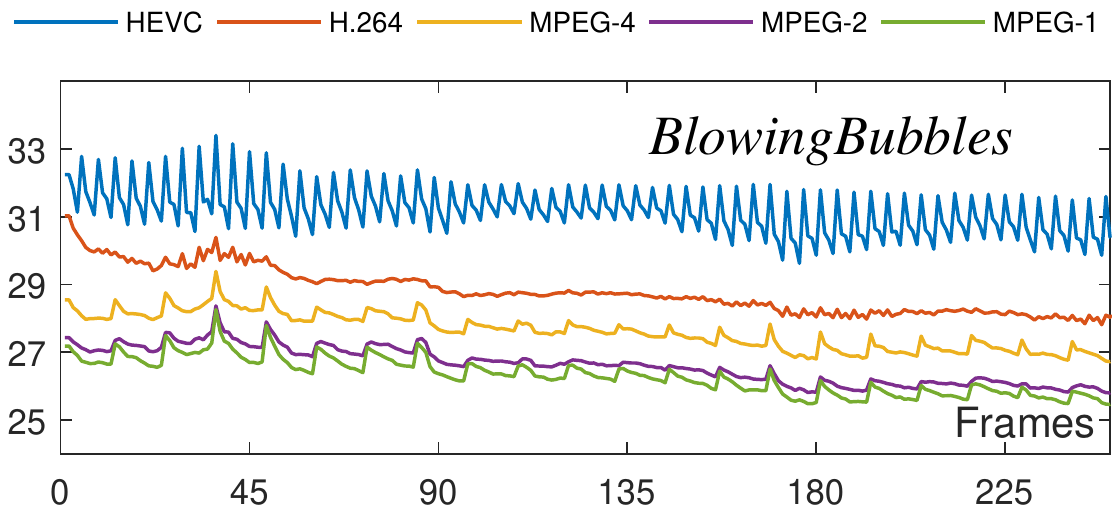}
	\end{minipage}
	
	\begin{minipage}[t]{2.3in}
		\centering
		\includegraphics[width = 2.3in]{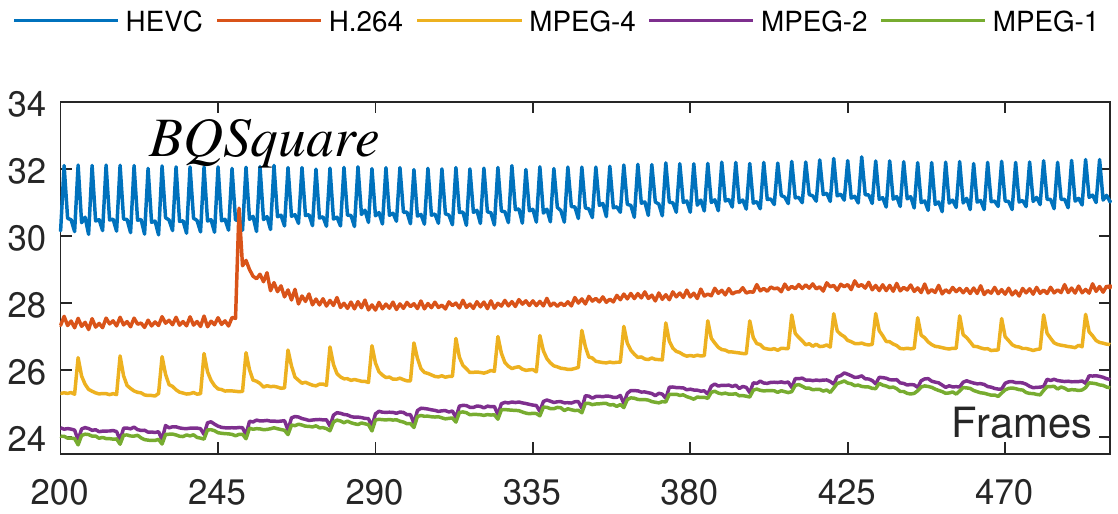}
	\end{minipage}
	\begin{minipage}[t]{2.3in}
		\centering
		\includegraphics[width = 2.3in]{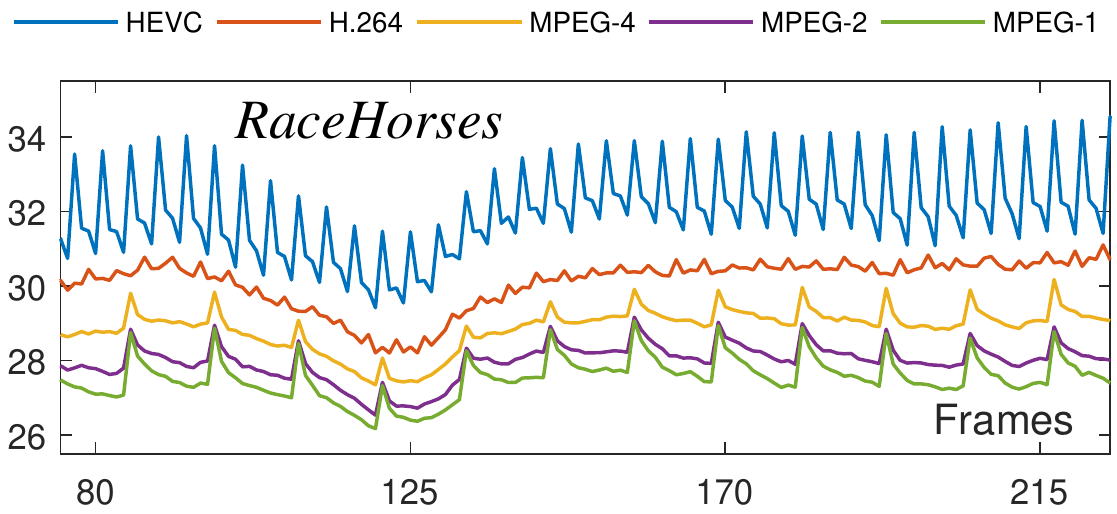}
	\end{minipage}
	\begin{minipage}[t]{2.3in}
		\centering
		\includegraphics[width = 2.3in]{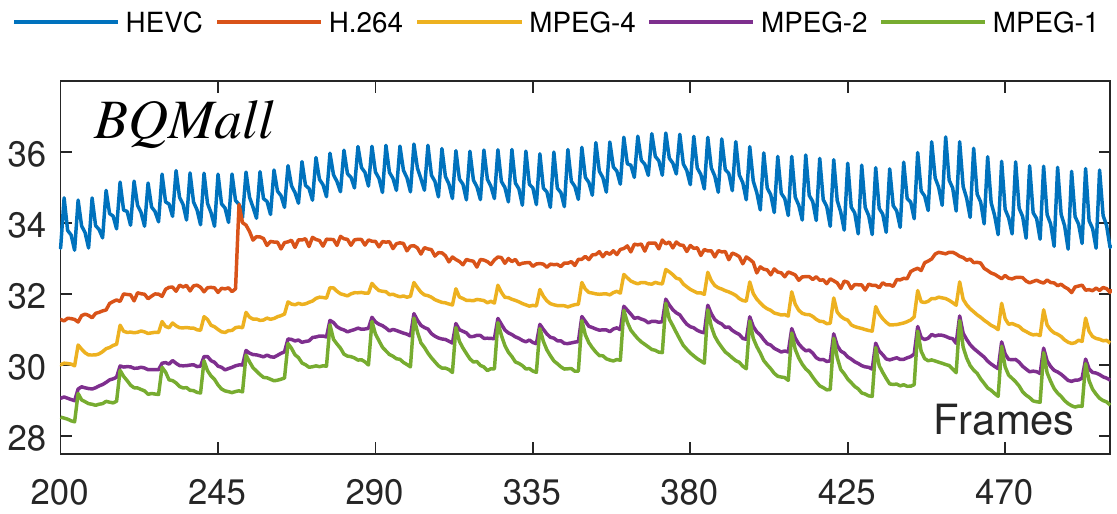}
	\end{minipage}
	\caption{PSNR (dB) curves of compressed video by various compression standards.}
	\label{PSNR_curve}
	\vspace{-1.0em}
\end{figure*}

\section{Related works}\label{works}

\subsection{Related works on quality enhancement}\label{works_QE}

Recently, extensive works  \cite{liew2004blocking,foi2007pointwise,wang2013adaptive,jancsary2012loss,jung2012image,chang2014reducing,dong2015compression,Guo2016Building, wang2016d3,Zhang2017Beyond,li2017efficient,Cavigelli2017CAS,tai2017memnet} have focused on enhancing the visual quality of compressed image.
Specifically, Foi \textit{et al.} \cite{foi2007pointwise} applied point-wise Shape-Adaptive DCT (SA-DCT) to reduce the blocking and ringing effects caused by JPEG compression.
Later, Jancsary \textit{et al.} \cite{jancsary2012loss} proposed reducing JPEG image blocking effects by adopting Regression Tree Fields (RTF).
Moreover, sparse coding was utilized to remove the JPEG artifacts, such as \cite{jung2012image} and \cite{chang2014reducing}.
Recently, deep learning has also been successfully applied to improve the visual quality of compressed images. Particularly, Dong \textit{et al.} \cite{dong2015compression} proposed a four-layer AR-CNN to reduce the JPEG artifacts of images.
Afterward, $\text{\textbf{D}}^3$ \cite{wang2016d3} and Deep Dual-domain Convolutional Network (DDCN) \cite{Guo2016Building} were proposed as advanced deep networks for the quality enhancement of JPEG image, utilizing the prior knowledge of JPEG compression.
Later, DnCNN was proposed in \cite{Zhang2017Beyond} for several tasks of image restoration, including quality enhancement. Li \textit{et al.} \cite{li2017efficient} proposed a 20-layer CNN for enhancing image quality.
Most recently, the memory network (MemNet) \cite{tai2017memnet} has been proposed for image restoration tasks, including quality enhancement.
In the MemNet, the memory block was introduced to generate the long-term memory across CNN layers, which successfully compensates the middle- and high-frequency signals distorted during compression.
It achieves the state-of-the-art quality enhancement performance for compressed images.

There are also some other works \cite{dai2017convolutional, yang2017decoder, Wang2017A} proposed for the quality enhancement of compressed video.
For example, the Variable-filter-size Residue-learning CNN (VRCNN) \cite{dai2017convolutional} was proposed to replace the in-loop filters for HEVC intra-coding.
However, the CNN in \cite{dai2017convolutional} was designed as a component of the video encoder, so that it is not practical for already compressed video.
Most recently, a Deep CNN-based Auto Decoder (DCAD), which contains 10 CNN layers, was proposed in \cite{Wang2017A} to reduce the distortion of compressed video.
Moreover, Yang \textit{et al.} \cite{yang2017decoder} proposed the DS-CNN approach for video quality enhancement.
In \cite{yang2017decoder}, DS-CNN-I and DS-CNN-B, as two subnetworks of DS-CNN, are used to reduce the artifacts of intra- and inter-coding, respectively.
All the above approaches can be seen as single-frame quality enhancement approaches, as they do not take any advantage of neighboring frames with high similarity.
Consequently, their performance on video quality enhancement is severely limited.

\subsection{Related works on multi-frame super-resolution}
\label{works_MFSR}

To our best knowledge, there exists no MFQE work for compressed video.
The closest area is multi-frame video super-resolution.
In the early years, Brandi \textit{et al.} \cite{brandi2008super} and Song \textit{et al.} \cite{song2011video} proposed to enlarge video resolution by taking advantage of high-resolution key frames.
Recently, many multi-frame super-resolution approaches have employed deep neural networks.
For example, Huang \textit{et al.} \cite{huang2018video} developed a Bidirectional Recurrent Convolutional Network (BRCN), which improves the super-resolution performance over traditional single-frame approaches.
Kappeler \textit{et al.} proposed a Video Super-Resolution network (VSRnet) \cite{Kappeler2016Video}, in which the neighboring frames are warped according to the estimated motion, and then both the current and warped neighboring frames are fed into a super-resolution CNN to enlarge the resolution of the current frame.
Later, Li \textit{et al.} \cite{Li2017Video} proposed replacing VSRnet by a deeper network with residual learning strategy.
All these multi-frame methods exceed the limitation of single-frame approaches (e.g., SR-CNN \cite{dong2016image}) for super-resolution, which only utilize the spatial information within one single frame.

Recently, the CNN-based FlowNet \cite{Dosovitskiy2015FlowNet,Ilg_2017_CVPR} has been applied in \cite{Makansi2017End} to estimate the motion across frames for super-resolution, which jointly trains the networks of FlowNet and super-resolution. Then, Caballero \textit{et al.} \cite{Caballero_2017_CVPR} designed a spatial transformer motion compensation network to detect the optical flow for warping neighboring frames. The current and warped neighboring frames were then fed into the Efficient Sub-Pixel Convolution Network (ESPCN) \cite{Shi2016Real} for super-resolution.
Most recently, the Sub-Pixel Motion Compensation (SPMC) layer has been proposed in \cite{Tao2017Detail} for video super-resolution. Besides, \cite{Tao2017Detail} utilized Convolutional Long Short-Term Memory (ConvLSTM) to achieve the state-of-the-art performance on video super-resolution.

The aforementioned multi-frame super-resolution approaches are motivated by the fact that different observations of a same object or scene are highly likely to exist in consecutive frames of video.
As a result, the neighboring frames may contain the content missed when down-sampling the current frame.
Similarly, for compressed video, the low-quality frames can be enhanced by taking advantage of their adjacent frames with higher quality, because heavy quality fluctuation exists across compressed frames.
Consequently, the quality of compressed videos may be effectively improved by leveraging the multi-frame information.
To the best of our knowledge, our MFQE approach proposed in this paper is the first attempt in this direction.

\section{Analysis of compressed video}
\label{quality}
In this section, we first establish a large-scale database of raw and compressed video sequences (Section\thinspace\ref{sec_database}) for training the deep neural networks in our MFQE approach.
We further analyze our database to investigate the frame-level quality fluctuation (Section\thinspace\ref{database_fluctuation}) and the similarity between consecutive compressed frames (Section\thinspace\ref{database_similairty}).
The analysis results can be seen as the motivation of our work.

\begin{figure*}[!t]
	\vspace{-1.0em}
	\centering
	\includegraphics[width = .9\linewidth]{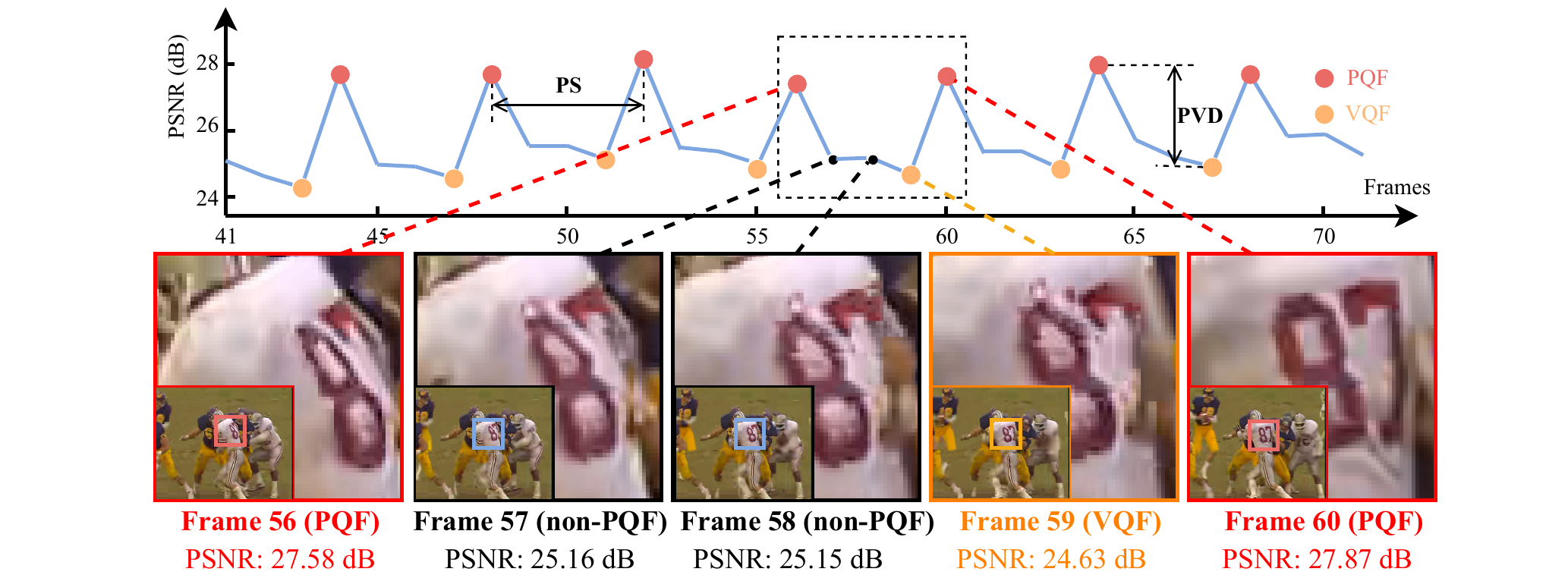}
	\caption{An example of frame-level quality fluctuation in video \textit{Football} compressed by HEVC.}
	\label{sq}
	\vspace{-0.0em}
\end{figure*}

\subsection{Database}
\label{sec_database}

First, we establish a database including 160 uncompressed video sequences.
These sequences are selected from the datasets of Xiph.org \cite{Xiph}, VQEG \cite{VQEG} and Joint Collaborative Team on Video Coding (JCT-VC) \cite{bossen2011common}.
The video sequences contained in our database are at large range of resolutions: SIF (352$\times$240), CIF (352$\times$288), NTSC (720$\times$486), 4CIF (704$\times$576), 240p (416$\times$240), 360p (640$\times$360), 480p (832$\times$480), 720p (1280$\times$720), 1080p (1920$\times$1080), and WQXGA (2560$\times$1600).
Moreover, Fig.\thinspace\ref{diversity} shows some typical examples of the sequences in our database, demonstrating the diversity of video content.
Then, all video sequences are compressed by MPEG-1 \cite{le1992mpeg}, MPEG-2 \cite{Schafer1995Digital}, MPEG-4 \cite{Sikora2002The}, H.264/AVC \cite{wiegand2003overview} and HEVC \cite{sullivan2012overview} at different quantization parameters (QPs)\footnote{FFmpeg is used for MPEG-1, MPEG-2, MPEG-4 and H.264/AVC compression, and HM16.5 is used for HEVC compression.}, to generate the corresponding video streams in our database.

\begin{figure}[!t]
	\vspace{-0em}
	\centering
	\includegraphics[width = 1\linewidth]{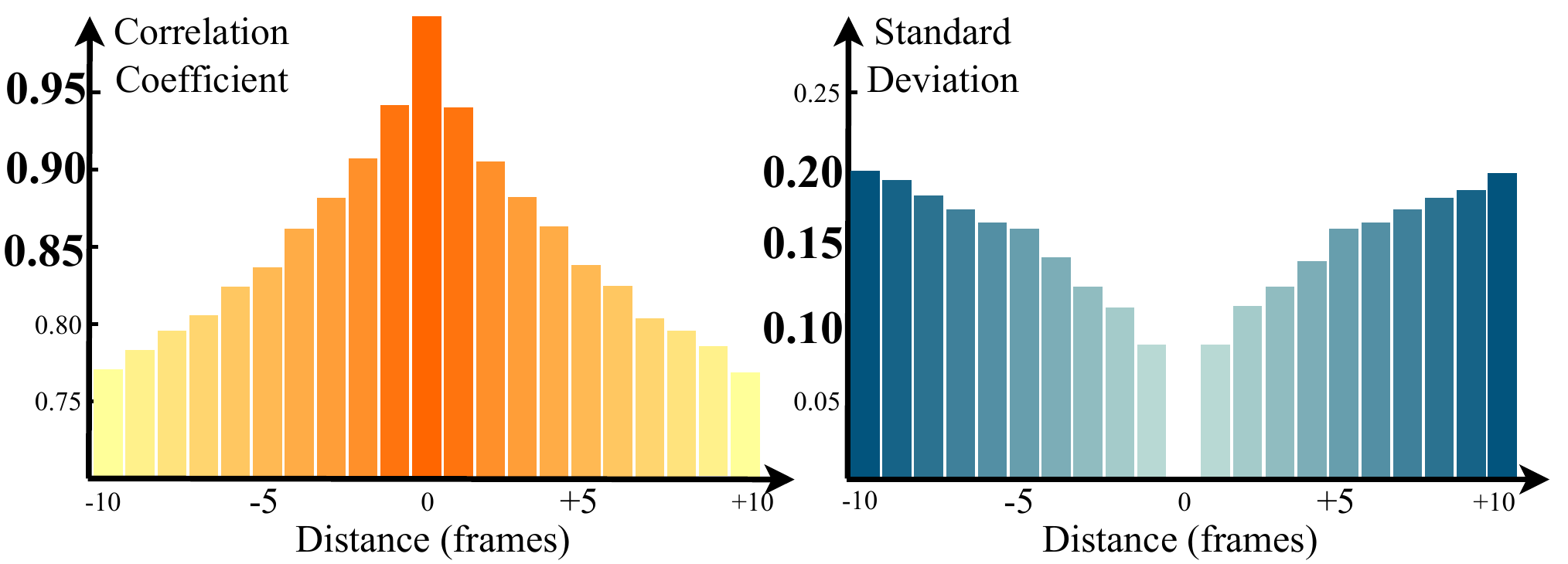}
	\caption{\footnotesize{The average CC value of each pair of adjacent frames in HEVC.}}
	\label{cc_HEVC}
	\vspace{-1.5em}
\end{figure}

\subsection{Frame-level quality fluctuation}
\label{database_fluctuation}

Fig.\thinspace\ref{PSNR_curve} shows the PSNR curves of 6 video sequences, which are compressed by different compression standards.
It can be seen that PSNR significantly fluctuates along with the compressed frames.
This indicates that there exists considerable quality fluctuation in compressed video sequences for MPEG-1, MPEG-2, MPEG-4, H.264/AVC and HEVC.
In addition, Fig.\thinspace\ref{sq} visualizes the subjective results of some frames in one video sequence, which is compressed by the latest HEVC standard.
We can see that visual quality varies across compressed frames, also implying the frame-level quality fluctuation.

Moreover, we measure the Standard Deviation (SD) of frame-level PSNR and Structural Similarity (SSIM) for each compressed video sequence, to quality fluctuation throughout the frames.
Besides, the Peak-Valley Difference (PVD), which calculates the average difference between peak values and their nearest valley values, is also measured for both PSNR and SSIM curves of each compressed sequence.
Note that the PVD reflects the quality difference between frames within a short period.
The results of SD and PVD are reported in Table\thinspace\ref{tab:psnr}, which are averaged over all 160 video sequences in our database.
Table\thinspace\ref{tab:psnr} shows that the average SD values of PSNR are above $0.87$ dB for all five compression standards.
This implies that compressed video sequences exist heavy fluctuation along with frames.
In addition, we can see from Table\thinspace\ref{tab:psnr} that the average PVD results of PSNR are above 1 dB for MPEG-1, MPEG-2, MPEG-4 and HEVC, except that of H.264 (0.4732 dB).
Therefore, the visual quality is dramatically different between PQFs and Valley Quality Frames (VQFs), such that it is possible to significantly improve the visual quality of VQFs given their neighboring PQFs.
Note that similar results can be found for SSIM as shown in Table\thinspace\ref{tab:psnr}.
In summary, we can conclude that the significant frame-level quality fluctuation exists for various video compression standards in terms of both PSNR and SSIM.

\begin{figure*}[!t]
	\vspace{-0.5em}
	\centering
	\includegraphics[width=.9\linewidth]{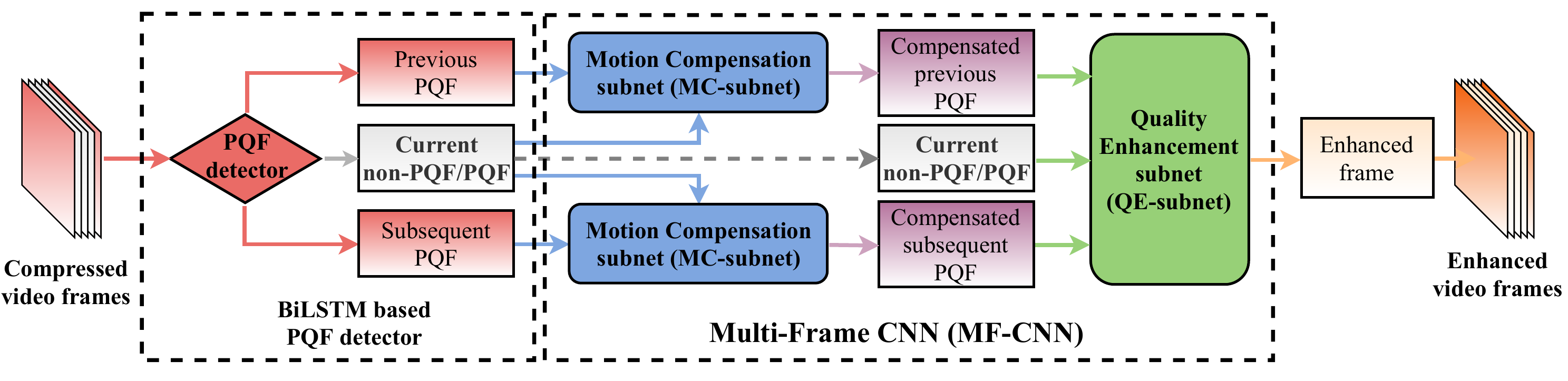}
	\caption{The framework of our proposed MFQE approach. Both non-PQFs and PQFs are enhanced by MF-CNN with the help of their nearest previous and subsequent PQFs. Note that the networks of enhancing PQFs and non-PQFs are trained, respectively.}
	\label{framework}
	\vspace{-0.5em}
\end{figure*}

\begin{table}[!t]
	\vspace{-0.5em}
	\centering
	\scriptsize
	\caption{Averaged SD, PVD and PS values of our database.}
	\begin{tabular}{l c c c c c}
		
		\toprule
		Metrics   & MPEG-1 & MPEG-2 & MPEG-4 & H.264 & HEVC \\
		
		\midrule
		\multicolumn{6}{c}{\textbf{PSNR (dB)}}\\
		
		\midrule
		SD & 2.2175 & 2.2273 & 2.1261 & 1.6899 & 0.8788 \\
		PVD & 1.1553 & 1.1665 & 1.0842 & 0.4732 & 1.1734 \\
		
		\midrule
		\multicolumn{6}{c}{\textbf{SSIM}}\\
		\midrule
		SD  & 0.0717 & 0.0726 & 0.0735 & 0.0552 & 0.0105 \\
		PVD & 0.0387 & 0.0391 & 0.0298 & 0.0102 & 0.0132 \\
		
		\midrule
		\multicolumn{6}{c}{\textbf{Separation (frames)}}\\
		\midrule
		PS & 5.3646 & 5.4713 & 5.4123 & 2.0529 & 2.6641 \\
		
		\bottomrule

	\end{tabular}
	\label{tab:psnr}
	\vspace{-0.0em}
\end{table}

\subsection{Similarity between neighboring frames}
\label{database_similairty}

It is intuitive that the frames within a short time period are with high similarity.
We thus evaluate the Correlation Coefficient (CC) values between each compressed frame and its previous/subsequent 10 frames, for all 160 sequences in our database.
The mean and SD of the CC values are shown in Fig.\thinspace\ref{cc_HEVC}, which are obtained from all sequences compressed by HEVC.
We can see that the average CC values are larger than 0.75 and the SD values of CC are less than 0.20, when the period of two frames is within 10. Similar results can be found for other four video compression standards.
This validates the high correlation of neighboring video frames.

In addition, it is necessary to investigate the number of non-PQFs between these two neighboring PQFs, denoted by the Peak Separation (PS), since the quality enhancement of each non-PQF is based on two neighboring PQFs.
Table\thinspace\ref{tab:psnr} also reports the results of PS, which are averaged over all 160 video sequences in our database.
We can see from this table\footnote{Note that this paper only defines PS according to PSNR rather than SSIM, but similar results can be found for SSIM.} that the PS values are considerably smaller than 10 frames, especially for the latest H.264 (PS = 2.0529) and HEVC (PS = 2.6641) standards.
Such a short distance, together with the similarity results in Fig.\thinspace\ref{cc_HEVC}, indicates the high similarity between two neighboring PQFs.
Therefore, the PQFs probably contain some useful content that is distorted in their neighboring non-PQFs.
Motivated by this, our MFQE approach is proposed to enhance the quality of non-PQFs through the advantageous information of the nearest PQFs.

\section{The proposed MFQE approach}
\label{MFQE}

\subsection{Framework}	

The framework of our MFQE approach is shown in Fig.\thinspace\ref{framework}.
As seen in this figure, our MFQE approach first detects PQFs that are used for quality enhancement of non-PQFs.
In practical application, raw sequences are not available in video quality enhancement, and thus PQFs and non-PQFs cannot be distinguished through comparison with raw sequences.
Therefore, we develop a no-reference PQF detector for our MFQE approach, which is detailed in Section\thinspace\ref{peakdetec}.
Then, we propose a novel MF-CNN architecture to enhance the quality of non-PQFs, which takes advantage of the nearest PQFs, i.e., both previous and subsequent PQFs.
As shown in Fig.\thinspace\ref{framework}, the MF-CNN architecture is composed of the MC-subnet and the QE-subnet. The MC-subnet (introduced in Section\thinspace\ref{mc}) is developed to compensate the temporal motion between neighboring frames.
To be specific, the MC-subnet firstly predicts the temporal motion between the current non-PQF and its nearest PQFs.
Then, the two nearest PQFs are warped with the spatial transformer according to the estimated motion.
As such, the temporal motion between non-PQF and PQFs can be compensated.
Finally, the QE-subnet (introduced in Section\thinspace\ref{cnn}), which has a spatio-temporal architecture, is proposed for quality enhancement.
In the QE-subnet, both the current non-PQF and compensated PQFs are the inputs, and then the quality of the non-PQF can be enhanced with the help of the adjacent compensated PQFs.
Note that, in the proposed MF-CNN, the MC-subnet and QE-subnet are trained jointly in an end-to-end manner.
Similarly, each PQF is also enhanced by MF-CNN with the help of its nearest PQFs.

\begin{figure}[!t]
	\vspace{-0.5em}
	\centering
	\includegraphics[width = 1\linewidth]{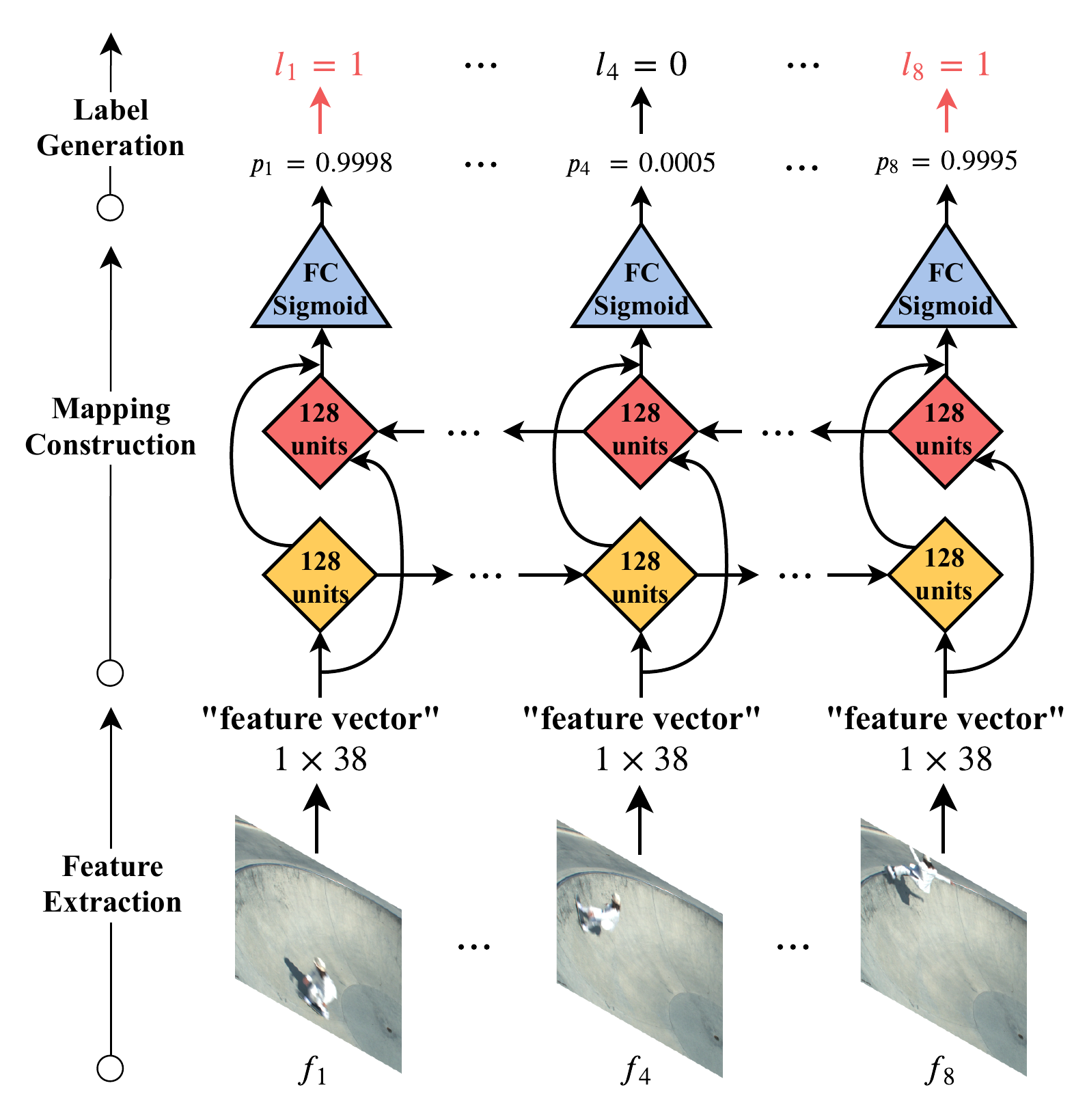}
	\caption{\footnotesize{The architecture of our BiLSTM based PQF detector.}}\label{BiLSTM-fig}
	\vspace{-1.0em}
\end{figure}

\subsection{BiLSTM-based PQF detector}
\label{peakdetec}

In our MFQE approach, the no-reference PQF detector is based on a BiLSTM network.
Recall that a PQF is the frame with higher quality than its adjacent frames.
Thus, the features of the current and neighboring frames in both forward and backward directions are used together to detect PQFs.
As revealed in Section\thinspace\ref{database_fluctuation}, the PQF frequently appears in compressed video, leading to the quality fluctuation.
Due to this, we apply the BiLSTM network \cite{Hochreiter2001A} as the PQF detector, in which the long- and short-term correlation between PQF and non-PQF can be extracted and modeled.

\vspace{0.3cm}
\noindent\textbf{Notations. }
We first introduce the notations for our PQF detector.
The consecutive frames in a compressed video are denoted by $\{f_n\}_{n=1}^{N}$, where $n$ indicates the frame order and $N$ is the total number of frames.
Then, the corresponding output from BiLSTM is denoted by $\{p_n\}_{n=1}^{N}$, in which $p_n$ is the probability of $f_n$ being a PQF.
Given $\{p_n\}_{n=1}^{N}$, the labels of PQFs for each frame can be determined and denoted by $\{l_n\}_{n=1}^{N}$. If $f_n$ is a PQF, then we have $l_n = 1$; otherwise, we have $l_n = 0$.

\vspace{0.3cm}
\noindent\textbf{Feature Extraction. }
Before training, we extract 38 features for each $f_n$.
Specifically, 2 compressed domain features, i.e., the number of assigned bits and quantization parameters, are extracted at each frame for detecting the PQF, since they are strongly related to visual quality and can be directly obtained from bitstream. In addition, we
follow  the no-reference quality assessment method \cite{seshadrinathan2010study} to extract 36 features at pixel domain.	
Finally, the extracted features are in form of a 38-dimension vector as the input to BiLSTM.

\vspace{0.3cm}
\noindent\textbf{Architecture. }
The architecture of the BiLSTM is shown in Fig.\thinspace\ref{BiLSTM-fig}.
As seen in this figure, the LSTM is bidirectional, in order to extract and model the dependencies from both forward and backward directions.
First, the input 38-dimension feature vector is fed into 2 LSTM cells, corresponding to either forward or backward direction.
Each of LSTM cells is composed of 128 units at one time step (corresponding to one video frame).
Then, the outputs of the bi-directional LSTM cells are fused and sent to the fully connected layer with a sigmoid activation.
Consequently, the fully connected layer outputs $p_n$,  as the probability of being the PQF frame.
Finally, the PQF label $l_n$ can be yielded upon $p_n$.

\begin{figure}[!t]
	\vspace{-0.5em}
	\centering
	\includegraphics[width = 1\linewidth]{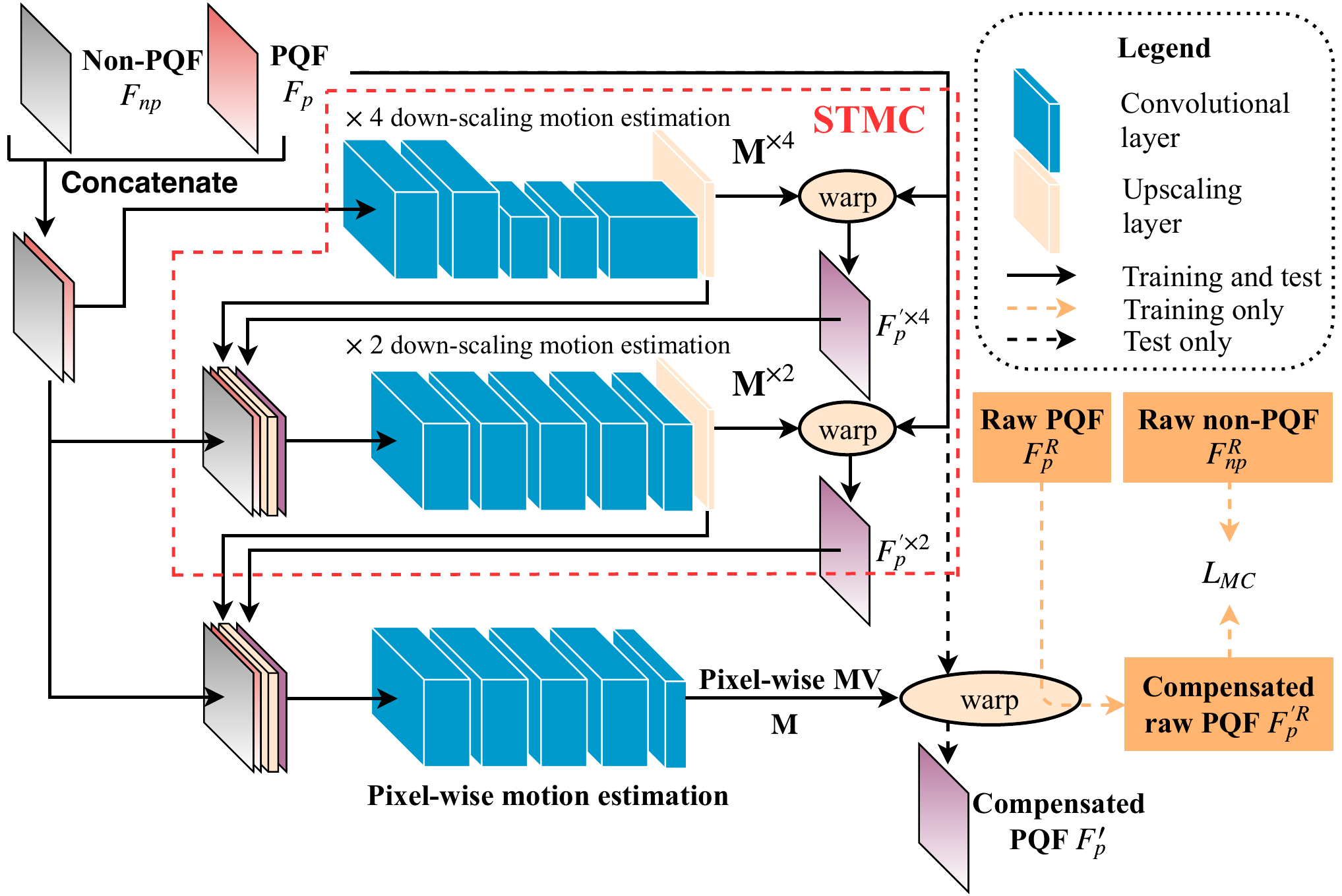}
	\caption{\footnotesize{The architecture of our MC-subnet.}}\label{mcn}
	\vspace{-1.0em}
\end{figure}

\vspace{0.3cm}
\noindent\textbf{Postprocessing.}
In our PQF detector, we further refine the results from BiLSTM according to the prior knowledge of PQF. Specifically, the following two strategies are developed to refine the labels $\{l_n\}_{n=1}^N$ of the PQF detector, where $N$ is the total number of frames.

\textit{Strategy I:}  Remove the consecutive PQFs.
According to the definition of PQF, it is impossible that the PQFs appear consecutively.
Hence, if the consecutive PQFs exist:
\begin{eqnarray}\label{c1}
\vspace{-1em}
\{l_{n+i}\}_{i=0}^j = 1\ \ \ \text{and}\ \ \ l_{n-1}=l_{n+j+1}=0,\ j\geq1,
\vspace{-1em}
\end{eqnarray}
we refine the PQF labels according to their probabilities:
\begin{eqnarray}\label{p1}
\vspace{-1em}
l_{n+i} = 0, \text{where}\ \ i \not= \mathop{\arg \max}_{0\leq k\leq j}(p_{n+k}),
\vspace{-1em}
\end{eqnarray}
so that only one PQF is left.

\textit{Strategy II}: Break the continuity of non-PQFs. According to the analysis in Section\thinspace\ref{quality}, PQFs frequently appear within a limited separation. For example, the average value of PS is 2.66 frames for HEVC compressed sequences. Here, we assume that $D$ is the maximal separation between two PQFs. Given this assumption, if the results of $\{l_n\}_{n=1}^N$ yield more than $D$ consecutive zeros (non-PQFs):
\begin{eqnarray}\label{c2}
\vspace{-1em}
\{l_{n+i}\}_{i=0}^d = 0\ \ \ \text{and}\ \ \ l_{n-1}=l_{n+d+1}=1, \ d > D,
\vspace{-1em}
\end{eqnarray}
then one of their corresponding frames $\{f_{n+i}\}_{i=0}^d$ need to act as a PQF. Accordingly, we set:
\begin{eqnarray}\label{p2}
\vspace{-1em}
l_{n+i} = 1, \text{where}\ \ i = \mathop{\arg\max}_{0 < k < d}(p_{n+k}).
\vspace{-1em}
\end{eqnarray}
After refining $\{l_n\}_{n=1}^N$ as discussed above, our PQF detector can locate PQFs and non-PQFs in the compressed video.

\subsection{MC-subnet}
\label{mc}

After detecting PQFs, our MFQE approach can enhance the quality of non-PQFs by taking advantage of their neighboring PQFs. Unfortunately, there exists considerable temporal motion between PQFs and non-PQFs. Hence, we develop the MC-subnet to compensate the temporal motion across frames, which is based on the CNN method of Spatial Transformer Motion Compensation  \cite{Caballero_2017_CVPR}.

\begin{table}[!t]
	\vspace{-0.5em}
	\renewcommand\arraystretch{1.1}
	\centering
	\footnotesize
	\caption{
		Convolutional layers for pixel-wise motion estimation.
	}
	\begin{tabular}{l c c c c c}
		\toprule
		Layers & Conv 1 & Conv 2 & Conv 3 & Conv 4 & Conv 5 \\
		\midrule
		Filter size & $3\times3$ & $3\times3$ & $3\times3$ & $3\times3$ & $3\times3$ \\

		Filter number & 24 & 24 & 24 & 24 & 2 \\

		Stride & 1 & 1& 1& 1& 1 \\

		Function &
		{PReLU} & {PReLU} & {PReLU} & {PReLU} & {Tanh} \\
		\bottomrule
	\end{tabular}
	\label{tab:conf1}
	\vspace{-1.0em}
\end{table}

\begin{figure*}[!t]
	\vspace{-0.5em}
	\centering
	\includegraphics[width = 1\linewidth]{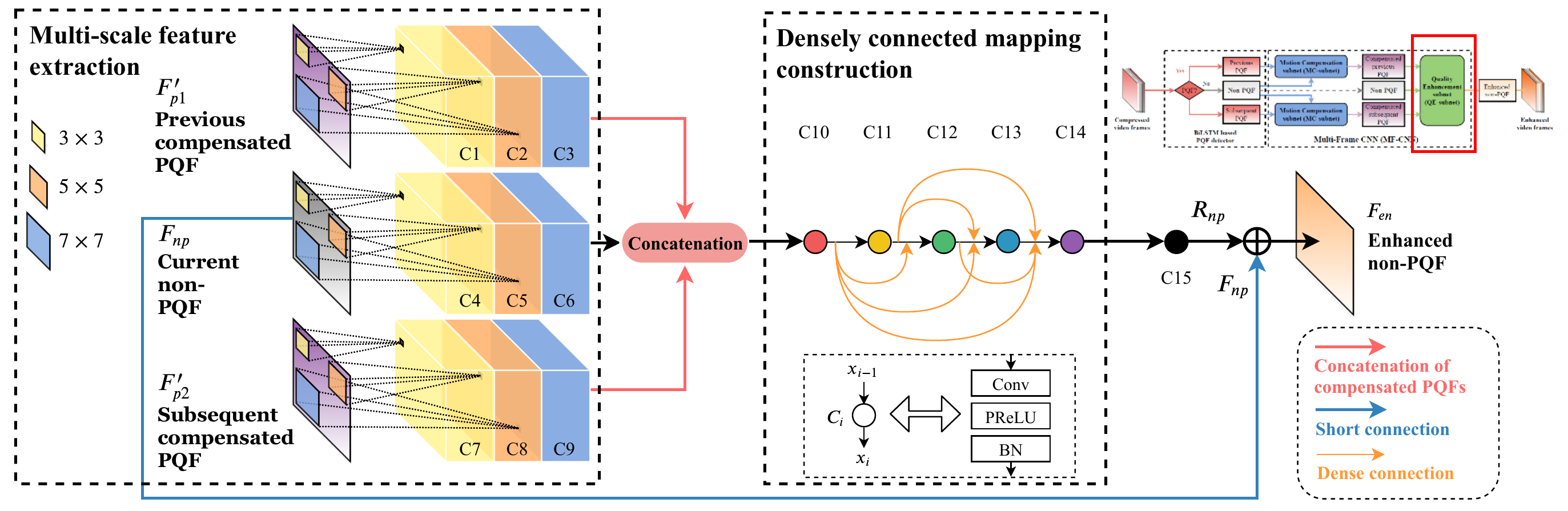}
	\caption{
			The architecture of our QE-subnet.
			In the multi-scale feature extraction component (denoted by C1-C9), the filter sizes of C1/4/7, C2/5/8 and C3/6/9 are $3 \times 3$, $5 \times 5$ and $7 \times 7$, respectively, and the filter number is set to 32 for each layer.
			Note that C1-C9 are directly applied to frames $F_{p1}^{'}$, $F_{np}$ or $F_{p2}^{'}$.
			In the densely connected mapping construction (denoted by C10-C14), the filter size and number are set to $3 \times 3$ and 32, respectively. The last layer C15 has only one filter with the size of $3 \times 3$. In addition, the PReLU activation is applied to C1-C14, while BN is applied to C10-C15.
	}
	\label{QE}
	\vspace{-0.5em}
\end{figure*}

\vspace{0.3cm}
\noindent\textbf{Architecture. }
The architecture of STMC is shown in Fig.\thinspace\ref{mcn}.
Additionally, the convolutional layers of pixel-wise motion estimation are described in Table\thinspace\ref{tab:conf1}.
The same as \cite{Caballero_2017_CVPR}, our MC-subnet adopts the convolutional layers to estimate the $\times4$ and $\times2$ down-scaling Motion Vector (MV) maps, denoted by $\mathbf{M}^{\times4}$ and $\mathbf{M}^{\times2}$. Down-scaling motion estimation is effective to handle large scale motion. However, because of down-scaling, the accuracy of MV estimation is reduced. Therefore, in addition to STMC, we further develop some additional convolutional layers for pixel-wise motion estimation in our MC-subnet, which does not contain any down-scaling process.
Then, the output of STMC includes the $\times2$ down-scaling MV map $\mathbf{M}^{\times2}$ and the corresponding compensated PQF $F'^{\times2}_{p}$. They are concatenated with the original PQF and non-PQF, as the input to the convolutional layers of the pixel-wise motion estimation.
Consequently, the pixel-wise MV map can be generated, which is denoted by $\mathbf{M}$. Note that the MV map $\mathbf{M}$ contains two channels, i.e., horizontal MV map $\mathbf{M}_x$ and vertical MV map $\mathbf{M}_y$. Here, $x$ and $y$ are the horizontal and vertical index of each pixel. Given $\mathbf{M}_x$ and $\mathbf{M}_y$, the PQF is warped to compensate the temporal motion. Let the compressed PQF and non-PQF be $F_p$ and $F_{np}$, respectively. The compensated PQF $F'_p$ can be expressed as
\begin{eqnarray}
\vspace{-.5em}
F'_p(x,y) = \mathcal{I}\{F_{p}(x+\mathbf{M}_x(x,y),y+\mathbf{M}_y(x,y))\},
\vspace{-.5em}
\end{eqnarray}
where $\mathcal{I}\{\cdot\}$ denotes bilinear interpolation. The reason for interpolation is that $\mathbf{M}_x(x,y)$ and $\mathbf{M}_y(x,y)$ may be non-integer values.

\vspace{0.3cm}
\noindent\textbf{Training strategy. }
Since it is hard to obtain the ground truth of MV, the parameters of the convolutional layers for motion estimation cannot be trained directly.
Instead, we can train the parameters by minimizing the MSE between the compensated adjacent frame and the current frame.
Note that the similar training strategy is adopted in \cite{Caballero_2017_CVPR} for motion compensation in video super-resolution tasks.
However, in our MC-subnet, both the input $F_p$ and $F_{np}$ are compressed frames with quality distortion.
Hence, when minimizing the MSE between $F'_p$ and the $F_{np}$, the MC-subnet learns to estimate the distorted MV, resulting in inaccurate motion estimation.
Therefore, the MC-subnet is trained under the supervision of the raw frames.
That is, we warp the raw frame of the PQF (denoted by $F^R_p$) using the MV map output from the convolutional layers of motion estimation, and minimize the MSE between the compensated raw PQF (denoted by $F'^{R}_{p}$) and the raw non-PQF (denoted by $F^R_{np}$).
Mathematically, the loss function of the MC-subnet can be written by
\begin{eqnarray}
\vspace{-1em}
L_{\text{MC}}(\theta_{mc})=||F'^{R}_p(\theta_{mc}) - F^R_{np}||_2^2,
\vspace{-1em}
\end{eqnarray}
where $\theta_{mc}$ represents the trainable parameters of our MC-subnet.
Note that the raw frames $F^R_p$ and $F^{R}_{np}$ are not required when compensating motion in test and practical use.

\subsection{QE-subnet}
\label{cnn}

Given the compensated PQFs, the quality of non-PQFs can be enhanced through the QE-subnet.
To be specific, the non-PQF $F_{np}$,  together with the compensated previous and subsequent PQFs ($F'_{p1}$ and $F'_{p2}$), are fed into the QE-subnet.
This way, both the spatial and temporal features of these three frames are extracted and fused,
such that the advantageous information in the adjacent PQFs can be used to enhance the quality of the non-PQF.
It differs from the conventional CNN-based single-frame quality enhancement approaches, which can only handle the spatial information within one single frame.

\vspace{0.3cm}
\noindent\textbf{Architecture. }
The architecture of QE-subnet is shown in Fig.\thinspace\ref{QE}.
The QE-subnet consists of two key lightweight components: multi-scale feature extraction (denoted by C1-9) and densely connected mapping construction (denoted by C10-14).

\begin{itemize}
	
\item Multi-scale feature extraction.
The input to the QE-subnet is non-PQF $F_{np}$ and its neighboring compensated PQFs $F_{p1}^{'}$ and $F_{p2}^{'}$.
Then, the spatial features of $F_{np}$, $F_{p1}^{'}$ and $F_{p2}^{'}$ are extracted by multi-scale convolutional filters, denoted by C1-9.
Specifically, the filter size of C1,4,7 is $3 \times 3$, while the filter sizes of C2,5,8 and C3,6,9 are $5 \times 5$ and $7 \times 7$, respectively.
The filter numbers of C1-9 are all 32.
After feature extraction, 288 feature maps filtered at different scales are obtained.
Subsequently, all feature maps from $F_{np}$, $F_{p1}^{'}$ and $F_{p2}^{'}$ are concatenated, and then flow into the dense connection component.

\item Densely connected mapping construction.
After obtaining the feature maps from $F_{np}$, $F_{p1}^{'}$ and $F_{p2}^{'}$, a densely connected architecture is applied to construct the non-linear mapping from feature maps to enhancement residual.
Note that enhancement residual refers to the difference between original and enhanced frames.
To be specific, there are 5 convolutional layers in the non-linear mapping of the densely connected architecture. Each of them has 32 convolutional filters with size of $3 \times 3$.
In addition, dense connection \cite{huang2017densely} is adopted to encourage feature reuse, strengthen feature propagation and mitigate the vanishing-gradient problem.
Moreover, Batch Normalization (BN) \cite{Ioffe2015Batch} is applied to all 5 layers after PReLU activation to reduce internal covariate shift, thus accelerating the training process.
We denote the composite non-linear mapping as $H_l(\cdot)$, including Convolution (Conv), PReLU and BN.
We further denote the output of the $l$-{th} layer as $x_l$, such that each layer can be formulated as follows,
\begin{equation}
\begin{split}
\vspace{-1em}
x_{11} &= H_{11}([x_{10}]) \\
x_{12} &= H_{12}([x_{10},x_{11}]) \\
x_{13} &= H_{13}([x_{10},x_{11},x_{12}]) \\
x_{14} &= H_{14}([x_{10},x_{11},x_{12},x_{13}]),
\vspace{-1em}
\end{split}
\end{equation}
where $[x_{10},x_{11},...,x_{14}]$ refers to the concatenation of the feature maps produced in layers C10-C14.
Finally, the enhanced non-PQF $F_{en}$ is generated by the pixel-wise summation of learned enhancement residual $R_{np}(\theta_{qe})$ and input non-PQF $F_{np}$
\begin{eqnarray}
F_{en} = F_{np} + R_{np}(\theta_{qe}),
\end{eqnarray}
where $\theta_{qe}$ is defined as the trainable parameters of the QE-subnet.

\end{itemize}

\vspace{0.3cm}
\noindent\textbf{Training strategy. }
The MC-subnet and QE-subnet in our MF-CNN are trained jointly in an end-to-end manner.
Recall that $F'^R_{p1}$ and $F'^R_{p2}$ are defined as the raw frames of the previous and incoming PQFs, respectively.
The loss function of our MF-CNN can be formulated as
\vspace{-.5em}
\begin{eqnarray}
\label{lossmf}
L_{\text{MF}}(\theta_{mc},\theta_{qe}) = a\cdot\underbrace{\sum_{i=1}^2||F'^R_{pi}(\theta_{mc})-F^R_{np}||_2^2}_{L_{\text{MC}}:\ \text{loss of MC-subnet}} \nonumber \\
+ b\cdot\underbrace{\big|\big|\big(F_{np}+R_{np}(\theta_{qe})\big)-F^{R}_{np}\big|\big|_2^2}_{L_{\text{QE}}:\ \text{loss of QE-subnet}}.
\end{eqnarray}
As \eqref{lossmf} indicates, the loss function of the MF-CNN is the weighted sum of $L_{\text{MC}}$ and $L_{\text{QE}}$, which are the $\ell_2$-norm training losses of MC-subnet and QE-subnet, respectively.
We divide the training into 2 steps.
In the first step, we set $a \gg b$, considering that $F'_{p1}$ and $F'_{p2}$ generated by MC-subnet are the basis of the following QE-subnet, and thus the convergence of MC-subnet is the primary target.
After the convergence of $L_{\text{MC}}$ is observed, we set $a\ll b$ to minimize the MSE between $F_{np}+R_{np}$ and $F^R_{np}$.
Finally, the MF-CNN model can be trained for video quality enhancement.

\begin{table}[!t]
	\vspace{-0.0em}
	\renewcommand\arraystretch{1.1}
	\centering
	\footnotesize
	\caption{Performance of our PQF detector on test sequences.}
	\begin{tabular}{|l|c||c c c|}
		\hline
		\multirow{2}{*}{Approach} & \multirow{2}{*}{QP} & Precision & Recall & $F_1$-score \\ [-0.3em]
		
		& & ($\%$) & ($\%$) & ($\%$) \\
		
		\hline
		\multirow{5}{*}{MFQE 2.0} & 22 & 100.0 & 95.9 & 97.8 \\
		
		& 27 & 98.2 & 94.1 & 96.1 \\
		
		& 32 & 100.0 & 84.3 & 90.7 \\
		
		\cline{2-5}
		& 37 & \textbf{100.0} & \textbf{96.5} & \textbf{98.2} \\
		
		& 42 & \textbf{100.0} & \textbf{97.3} & \textbf{98.6} \\
		
		\hline
		\multirow{2}{*}{MFQE 1.0} & 37 & 90.7 & 92.1 & 91.1 \\
		
		& 42 & 94.0 & 90.9 & 92.2\\
		
		\hline
		
	\end{tabular}
	\label{sc}
	\vspace{-0.0em}
\end{table}

\begin{figure*}[!t]
	
	\centering
	\includegraphics[width = 1\linewidth]{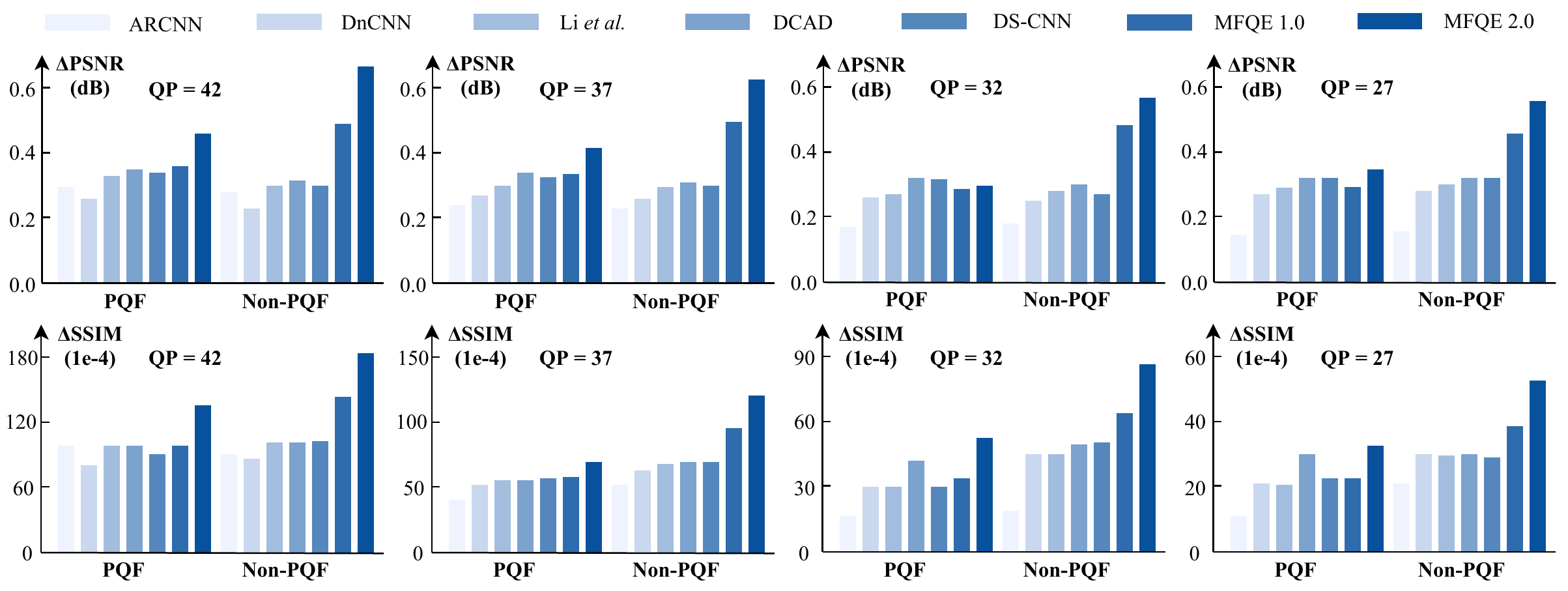}
	\caption{
		Average results of $\Delta$PSNR (dB) and $\Delta$SSIM for PQFs and non-PQFs in all test sequences at different QPs.
	}
	\label{deltapsnr}
	\vspace{-0.0em}
\end{figure*}

\section{Experiments}
\label{exp}

\subsection{Settings}
\label{setting}

In this section, the experimental results are presented to validate the effectiveness of our MFQE 2.0 approach.
Note that our MFQE 2.0 approach is called MFQE in this paper, while the MFQE approach of our conference paper \cite{yang2018multi} is named as MFQE 1.0 for comparison.
In our database, except for 18 standard test sequences of Joint Collaborative Team on Video Coding (JCT-VC)\cite{ohm2012comparison}, other 142 sequences are randomly divided into non-overlapping training set (106 sequences) and validation set (36 sequences).
We compress all 160 sequences by HM16.5 under Low-Delay configuration, setting the Quantization Parameters (QPs) to 22, 27, 32, 37 and 42, respectively.

For the BiLSTM-based PQF detector, the hyper-parameter $D$ of \eqref{c2} is set to 3 in post-processing\footnote{$D$ should be adjusted according to the compression standard and configuration.},  because the average value of PS is 2.66 frames for HEVC compressed sequences.
In addition, the LSTM length is set to 8.
Before training the MF-CNN, the raw and compressed sequences are segmented into $64\times64$ patches as the training samples.
The batch size is set to be 128.
We apply the Adam algorithm \cite{Kingma2014Adam} with the initial learning rate as $10^{-4}$ to minimize the loss function \eqref{lossmf}.
It is worth mentioning that the MC-subnet may be unable to converge, if the initial learning rate is oversize, e.g., $10^{-3}$.
For QE subnet, we set $a = 1$ and $b = 0.01$ in \eqref{lossmf} at first to make the MC-subnet convergent.
After the convergence, we set $a = 0.01$ and $b = 1$, so that the QE-subnet can converge faster.

\begin{table}[t]
	\vspace{-0.0em}
	\renewcommand\arraystretch{1.1}
	\centering
	\footnotesize
	\caption{Performance of our PQF detector on test sequences at QP = 37.}
	\begin{tabular}{|c|l||c c c|}
		\hline
		\multicolumn{2}{|l||}{\multirow{2}{*}{Sequence}} &  Precision & Recall & $F_1$-score \\[-0.3em]
		
		\multicolumn{2}{|l||}{} & ($\%$) & ($\%$) & ($\%$) \\
		
		\hline
		\multirow{2}{*}[-0em]{A} & \textit{Traffic}  & 100.0 & 97.4 & 98.7 \\
		
		& \textit{PeopleOnStreet}  & 100.0 & 97.4 & 98.7 \\
		
		\hline
		\multirow{5}{*}[-0em]{B} & \textit{Kimono} & 100.0 & 98.4 & 99.2 \\
		
		& \textit{ParkScene} & 100.0 & 98.4 & 99.2 \\
		
		& \textit{Cactus} & 100.0 & 99.2 & 99.6 \\
		
		& \textit{BQTerrace}  & 100.0 & 96.2 & 98.0 \\
		
		& \textit{BasketballDrive}  & 100.0 & 97.4 & 98.7 \\
		
		\hline
		\multirow{4}{*}[-0em]{C} & \textit{RaceHorses}  & 100.0 & 93.8 & 96.8 \\
		
		& \textit{BQMall}  & 100.0 & 98.7 & 99.3 \\
		
		& \textit{PartyScene} & 100.0 & 98.4 & 99.2 \\

		& \textit{BasketballDrill} & 100.0 & 91.9 & 95.8 \\
		
		\hline
		\multirow{4}{*}[-0em]{D} & \textit{RaceHorses}  & 100.0 & 94.9 & 97.4 \\

		& \textit{BQSquare} & 100.0 & 86.2 & 92.6 \\

		& \textit{BlowingBubbles}  & 100.0 & 98.4 & 99.2 \\

		& \textit{BasketballPass}  & 100.0 & 94.0 & 96.9 \\
		
		\hline
		\multirow{3}{*}[-0em]{E} & \textit{FourPeople} & 100.0 & 99.3 & 99.7 \\

		& \textit{Johnny}  & 100.0 & 98.0 & 99.0 \\

		& \textit{KristenAndSara} & 100.0 & 99.3 & 99.7 \\
		
		\hline
		\multicolumn{2}{|c||}{Average}  & \textbf{100.0} & \textbf{96.5} & \textbf{98.2} \\
		
		\hline	
		
	\end{tabular}
	\label{sc2}
	\vspace{-0.5em}
\end{table}

\subsection{Performance of the PQF detector}
The performance of PQF detection is critical, since it is the first process of our MFQE approach.
Thus, we evaluate the performance of our BiLSTM-based approach in PQF detection.
For evaluation, we measure precision, recall and $F_1$-score of PQF detection over all 18 test sequences compressed at five QPs (= 22, 27, 32, 37 and 42).
The average results are shown in Table\thinspace\ref{sc}.
In this table, we also list the results of PQF detection by  the SVM-based approach of MFQE 1.0 as reported in \cite{yang2018multi}.
Note that the results of only two QPs (= 37 and 42) are reported in \cite{yang2018multi}.

We can see from Table\thinspace\ref{sc} that the proposed BiLSTM-based PQF detector in MFQE 2.0 performs well in terms of precision, recall and  $F_1$-score.
For example, at QP = 37, the average precision, recall and $F_1$-score of our BiLSTM-based PQF detector are  $100.0\%$, $96.5\%$ and $98.2\%$, considerably higher than those of the SVM-based approach in MFQE 1.0.
More importantly, the PQF detection of our approach is robust to all 5 QPs, since the average values of $F_1$-score are all above $90\%$.
In addition, Table\thinspace\ref{sc2} shows the performance of our BiLSTM-based PQF detector over each of 18 test sequences compressed at QP = 37.
As seen in this table, the high performance is achieved by our PQF detector for almost all sequences, as only the recall of sequence \textit{BQSquare} is below $90\%$.
In conclusion, the effectiveness of our BiLSTM-based PQF detector is validated, laying a firm foundation for our MFQE approach.

\subsection{Performance of our MFQE approach}
\label{MFQEperformance}

In this section, we evaluate the quality enhancement performance of our MFQE approach in terms of $\Delta$PSNR, which measures the PSNR gap between the enhanced and original compressed sequences.
In addition, the structural similarity (SSIM) index is also evaluated.
Then, the performance of our MFQE approach is compared with those of AR-CNN \cite{dong2015compression}, DnCNN \cite{Zhang2017Beyond}, Li \textit{et al.} \cite{li2017efficient}, DCAD \cite{Wang2017A} and DS-CNN \cite{yang2018enhancing}.
Among them, AR-CNN, DnCNN and Li \textit{et al.} are the latest quality enhancement approaches for compressed images, while DCAD and DS-CNN are the state-of-the-art video quality enhancement approaches.
For fair comparison, all compared approaches are retrained over our training set, the same as our MFQE approach.

\vspace{0.3cm}
\noindent\textbf{Quality enhancement on non-PQFs. }
Our MFQE approach mainly focuses on enhancing the quality of non-PQFs using the neighboring multi-frame information.
Therefore, we first assess the quality enhancement of non-PQFs.
Fig.\thinspace\ref{deltapsnr} shows the $\Delta$PSNR and $\Delta$SSIM results averaged over PQFs and non-PQFs of all 18 test sequences compressed at 4 different QPs.
As shown, our MFQE approach significantly outperforms other approaches on non-PQF enhancement.
The average improvement of non-PQF quality is 0.614 dB and 0.012 in SSIM, while that of the second-best approach is 0.317 dB in PSNR and 0.007 in SSIM.
We can further see from Fig.\thinspace\ref{deltapsnr} that our MFQE approach has a considerably larger PSNR improvement for non-PQFs, compared to that for PQFs.
By contrast, for compared approaches, the PSNR improvement of non-PQFs is similar to or even less than that of PQFs.
In a word, the above results validate the outstanding effectiveness of our MFQE approach in enhancing the quality of non-PQFs.

\begin{table*}[t]
	\vspace{-0.0em}
	\renewcommand\arraystretch{1.3}
	\centering
	\scriptsize
	\caption{Overall comparison for $\Delta$PSNR (dB) and $\Delta$SSIM ($\times10^{-4}$) over test sequences at five QPs.}
	\begin{threeparttable}
		
		\begin{tabular}{|c|c|l||c c c c c c c c c c c c c c|}
			
			\hline
			\multirow{2}{*}{QP} & \multicolumn{2}{c||}{\multirow{2}{*}{Approach}} & \multicolumn{2}{c}{AR-CNN} & \multicolumn{2}{c}{DnCNN} & \multicolumn{2}{c}{Li \textit{et al.}} & \multicolumn{2}{c}{DCAD} & \multicolumn{2}{c}{DS-CNN} & \multicolumn{2}{c}{\multirow{2}{*}{MFQE 1.0}} & \multicolumn{2}{c|}{\multirow{2}{*}{MFQE 2.0}} \\ [-0.3em]
			
			& \multicolumn{2}{c||}{} & \multicolumn{2}{c}{\cite{dong2015compression}\tnote{*}}  & \multicolumn{2}{c}{\cite{Zhang2017Beyond}} & \multicolumn{2}{c}{\cite{li2017efficient}} & \multicolumn{2}{c}{\cite{Wang2017A}} & \multicolumn{2}{c}{\cite{yang2018enhancing}} & \multicolumn{2}{c}{} & \multicolumn{2}{c|}{} \\
			
			\hline
			\multirow{20}{*}{37} & \multicolumn{2}{c||}{Metrics} & \tiny PSNR & \tiny SSIM & \tiny PSNR & \tiny SSIM & \tiny PSNR & \tiny SSIM & \tiny PSNR & \tiny SSIM & \tiny PSNR & \tiny SSIM & \tiny PSNR & \tiny SSIM & \tiny PSNR & \tiny SSIM \\
			
			\cline{2-17}
			& \multirow{2}{*}{A} & \textit{Traffic}
			& 0.239 & 47 & 0.238 & 57 & 0.293 & 60 & 0.308 & 67 & 0.286 & 60 & 0.497 & 90 & \textbf{0.585} & \textbf{102}\\
			
			& & \textit{PeopleOnStreet}
			& 0.346 & 75 & 0.414 & 82 & 0.481 & 92 & 0.500 & 95 & 0.416 & 85 & 0.802 & 137 & \textbf{0.920} & \textbf{157} \\
			
			\cline{2-17}
			& \multirow{5}{*}{B} & \textit{Kimono}
			& 0.219 & 65 & 0.244 & 75 & 0.279 & 78 & 0.276 & 78 & 0.249 & 75 & 0.495 & 113 & \textbf{0.550} & \textbf{118} \\
			
			& & \textit{ParkScene}
			& 0.136 & 38 & 0.141 & 50 & 0.150 & 48 & 0.160 & 50 & 0.153 & 50 & 0.391 & 103 & \textbf{0.457} & \textbf{123}\\
			
			& & \textit{Cactus}
			& 0.190 & 38 & 0.195 & 48 & 0.232 & 58 & 0.263 & 58 & 0.239 & 58 & 0.439 & 88 & \textbf{0.501} & \textbf{100}\\
			
			& & \textit{BQTerrace}
			& 0.195 & 28 & 0.201 & 38 & 0.249 & 48 & 0.279 & 50 & 0.257 & 48 & 0.270 & 48 & \textbf{0.403} & \textbf{67}\\
			
			& & \textit{BasketballDrive}
			& 0.229 & 55 & 0.251 & 58 & 0.296 & 68 & 0.305 & 68 & 0.282 & 65 & 0.406 & 80 & \textbf{0.465} & \textbf{83}\\
			
			\cline{2-17}
			& \multirow{4}{*}{C} & \textit{RaceHorses}
			& 0.219 & 43 & 0.253 & 65 & 0.276 & 65 & 0.282 & 65 & 0.267 & 63 & 0.340 & 55 & \textbf{0.394} & \textbf{80}\\
			
			& & \textit{BQMall}
			& 0.275 & 68 & 0.281 & 68 & 0.325 & 88 & 0.340 & 88 & 0.330 & 80 & 0.507 & 103 & \textbf{0.618} & \textbf{120}\\
			
			& & \textit{PartyScene}
			& 0.107 & 38 & 0.131 & 48 & 0.131 & 45 & 0.164 & 48 & 0.174 & 58 & 0.217 & 73 &\textbf{0.363} & \textbf{118}\\
			
			& & \textit{BasketballDrill}
			& 0.247 & 58 & 0.331 & 68 & 0.376 & 88 & 0.386 & 78 & 0.352 & 68 & 0.477 & 90 &\textbf{0.579} &\textbf{120}\\
			
			\cline{2-17}
			& \multirow{4}{*}{D} & \textit{RaceHorses}
			& 0.268 & 55 & 0.311 & 73 & 0.328 & 83 & 0.338 & 83 & 0.318 & 75 & 0.507 & 113 & \textbf{0.594} & \textbf{143}\\
			
			& & \textit{BQSquare}
			& 0.080 & 8 & 0.129 & 18 & 0.086 & 25 & 0.197 & 38 & 0.201 & 38 & -0.010 & 15 & \textbf{0.337} & \textbf{65}\\
			
			& & \textit{BlowingBubbles}
			& 0.164 & 35 & 0.184 & 58 & 0.207 & 68 & 0.215 & 65 & 0.228 & 68 & 0.386 & 120 & \textbf{0.533} & \textbf{170}\\
			
			& & \textit{BasketballPass}
			& 0.259 & 58 & 0.307 & 75 & 0.343 & 85 & 0.352 & 85 & 0.335 & 78 & 0.628 & 138 & \textbf{0.728} & \textbf{155}\\
			
			\cline{2-17}
			& \multirow{3}{*}{E} & \textit{FourPeople}
			& 0.373 & 50 & 0.388 & 60 & 0.449 & 70 & 0.506 & 78 & 0.459 & 70 & 0.664 & 85 & \textbf{0.734} & \textbf{95}\\
			
			& & \textit{Johnny}
			& 0.247 & 10 & 0.315 & 40 & 0.398 & 60 & 0.410 & 50 & 0.378 & 40 & 0.548 & 55 & \textbf{0.604} & \textbf{68}\\
			
			& & \textit{KristenAndSara}
			& 0.409 & 50 & 0.421 & 60 & 0.485 & 68 & 0.524 & 70 & 0.481 & 60 & 0.655 & 75 & \textbf{0.754} & \textbf{85}\\
			
			\cline{2-17}
			& \multicolumn{2}{c||}{Average}
			& 0.233 & 45 & 0.263 & 58 & 0.299 & 66 & 0.322 & 67 & 0.300 & 63 & 0.455 & 88 & \textbf{0.562} & \textbf{109}\\
			
			\hline
			\hline
			42 & \multicolumn{2}{c||}{Average}
			& 0.285 & 96 & 0.221 & 77 & 0.318 & 105 & 0.324 & 109 & 0.310 & 101 & 0.444 & 130 & \textbf{0.589} & \textbf{165}\\
			
			\hline
			32 & \multicolumn{2}{c||}{Average}
			& 0.176 & 19 & 0.256 & 35 & 0.275 & 37 & 0.316 & 44 & 0.273 & 38 & 0.431 & 58 & \textbf{0.516} & \textbf{68}\\
			
			\hline
			27 & \multicolumn{2}{c||}{Average}
			& 0.177 & 14 & 0.272 & 24 & 0.295 & 28 & 0.316 & 30 & 0.267 & 23 & 0.399 & 34 & \textbf{0.486} & \textbf{42} \\
			
			\hline
			22 & \multicolumn{2}{c||}{Average}
			& 0.142 & 8 & 0.287 & 18 & 0.300 & 19 & 0.313 & 19 & 0.254 & 15 & 0.307 & 19 & \textbf{0.458} & \textbf{27}\\
			\hline
			
		\end{tabular}

		\begin{tablenotes}
			\footnotesize
			\item[*] All compared approaches in this paper are retrained over our training set, the same as MFQE 2.0.
		\end{tablenotes}

		\label{dpsnr}
	\end{threeparttable}
	\vspace{+0em}
\end{table*}

\begin{table*}[!t]
	\vspace{+0.0em}
	\renewcommand\arraystretch{1.3}
	\centering
	\scriptsize
	\caption{Overall BD-BR reduction ($\%$) of test sequences with the HEVC baseline as an anchor. \protect\\Calculated at QP = 22, 27, 32, 37 and 42.}
	
	\begin{tabular}{|c|l|| c c c c c c c|}
		
		\hline
		\multicolumn{2}{|c||}{Sequence} & AR-CNN & DnCNN & Li \textit{et al.} & DCAD & DS-CNN & MFQE 1.0 & MFQE 2.0 \\
		
		\hline
		\multirow{2}{*}{A} & \textit{Traffic}
		& 7.40 & 8.54 & 10.08 & 9.97 & 9.18 & 14.56 & \textbf{16.98} \\
		
		& \textit{PeopleOnStreet}
		& 6.99 & 8.28 & 9.64 & 9.68 & 8.67 & 13.71 & \textbf{15.08} \\
		
		\cline{1-9}
		\multirow{5}{*}{B} & \textit{Kimono}
		& 6.07 & 7.33 & 8.51 & 8.44 & 7.81 & 12.60 & \textbf{13.34} \\
		
		& \textit{ParkScene}
		& 4.47 & 5.04 & 5.35 & 5.68 & 5.42 & 12.04 & \textbf{13.66} \\
		
		& \textit{Cactus}
		& 6.16 & 6.80 & 8.23 & 8.69 & 8.78 & 12.78 & \textbf{14.84} \\
		
		& \textit{BQTerrace}
		& 6.86 & 7.62 & 8.79 & 9.98 & 8.67 & 10.95 & \textbf{14.72} \\
		
		& \textit{BasketballDrive}
		& 5.83 & 7.33 & 8.61 & 8.94 & 7.89 & 10.54 & \textbf{11.85} \\
		
		\cline{1-9}
		\multirow{4}{*}{C} & \textit{RaceHorses}
		& 5.07 & 6.77 & 7.10 & 7.62 & 7.48 &8.83& \textbf{9.61} \\
		
		& \textit{BQMall}
		& 5.60 & 7.01 & 7.79 & 8.65 & 7.64 & 11.11 & \textbf{13.50} \\
		
		& \textit{PartyScene}
		& 1.88 & 4.02 & 3.78 & 4.88 & 4.08 & 6.67 & \textbf{11.28} \\
		
		& \textit{BasketballDrill}
		& 4.67 & 8.02 & 8.66 & 9.80 & 8.22 & 10.47 & \textbf{12.63} \\
		
		\cline{1-9}
		\multirow{4}{*}{D} & \textit{RaceHorses}
		& 5.61 & 7.22 & 7.68 & 8.16 & 7.35 & 10.41 & \textbf{11.55} \\
		
		& \textit{BQSquare}
		& 0.68 & 4.59 & 3.59 & 6.11 & 3.94 & 2.72 & \textbf{11.00} \\
		
		& \textit{BlowingBubbles}
		& 3.19 & 5.10 & 5.41 & 6.13 & 5.55 & 10.73 & \textbf{15.20} \\
		
		& \textit{BasketballPass}
		& 5.11 & 7.03 & 7.78 & 8.35 & 7.49 & 11.70 & \textbf{13.43} \\
		
		\cline{1-9}
		\multirow{3}{*}{E} & \textit{FourPeople}
		& 8.42 & 10.12 & 11.46 & 12.21 & 11.13 & 14.89 & \textbf{17.50} \\
		
		& \textit{Johnny}
		& 7.66 & 10.91 & 13.05 & 13.71 & 12.19 & 15.94 & \textbf{18.57} \\
		
		& \textit{KristenAndSara}
		& 8.94 & 10.65 & 12.04 & 12.93 & 11.49 & 15.06 & \textbf{18.34} \\
		
		\cline{1-9}
		\multicolumn{2}{|c||}{Average} & 5.59 & 7.36 & 8.20 & 8.89 & 7.85 & 11.41 & \textbf{14.06} \\
		
		\hline
		
	\end{tabular}
	\label{tab_bdr}
	\vspace{+0em}
\end{table*}

\begin{figure}[!t]
	\vspace{-0.0em}
	\centering
	\begin{minipage}[t]{1.73in}
		\centering
		\includegraphics[width=1.73in]{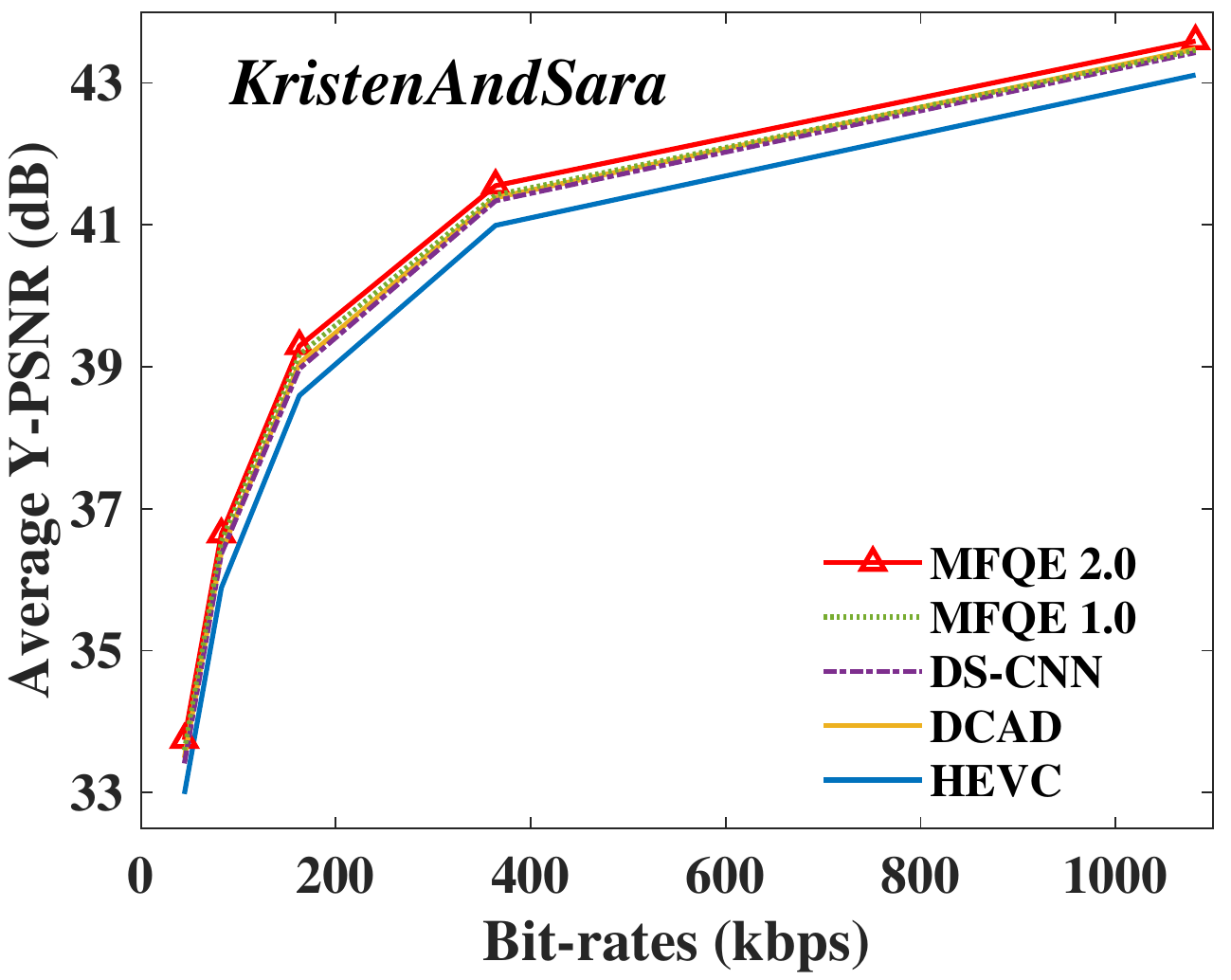}
	\end{minipage}
	\begin{minipage}[t]{1.73in}
		\centering
		\includegraphics[width=1.73in]{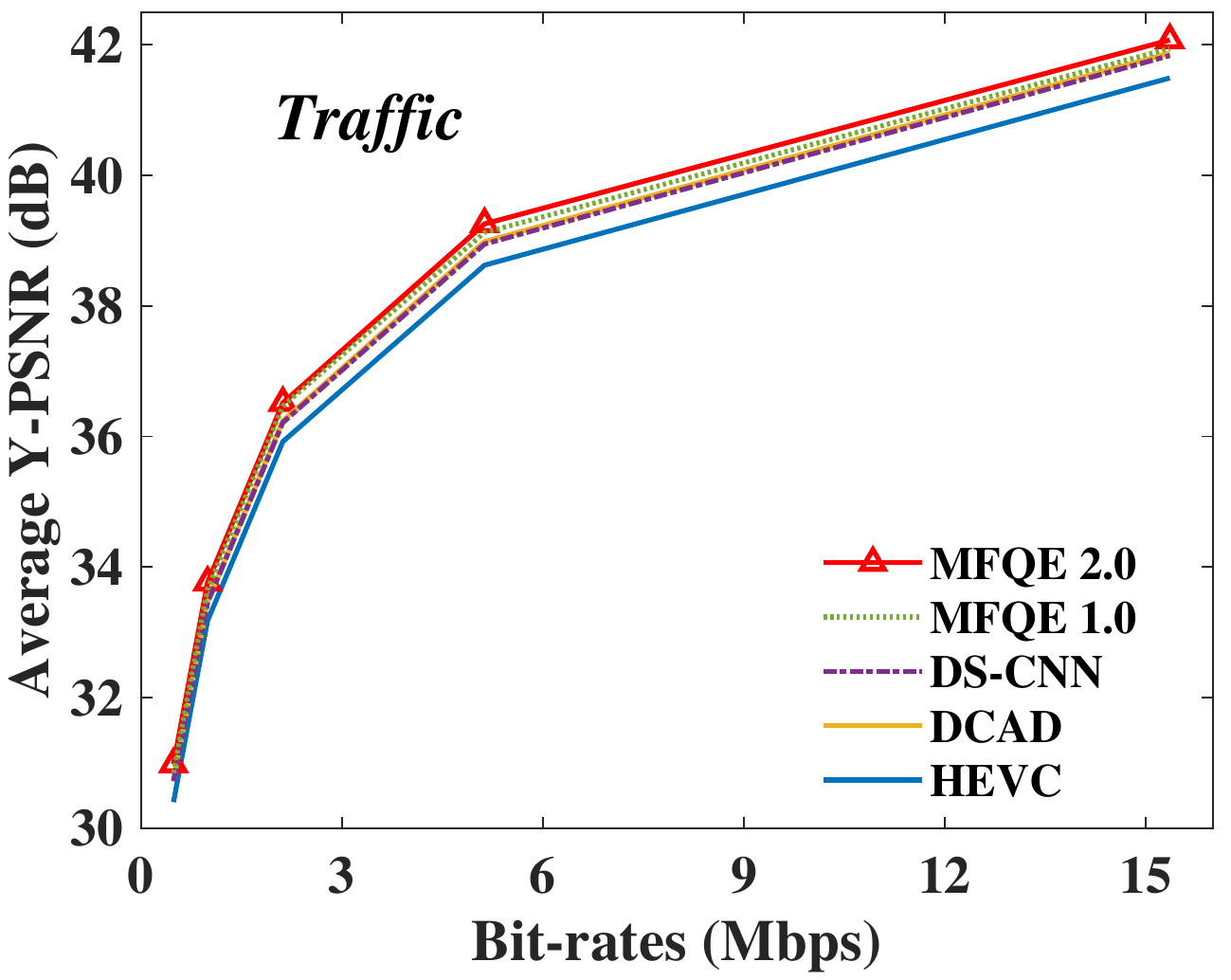}
	\end{minipage}
	
	\begin{minipage}[t]{1.73in}
		\centering
		\includegraphics[width=1.73in]{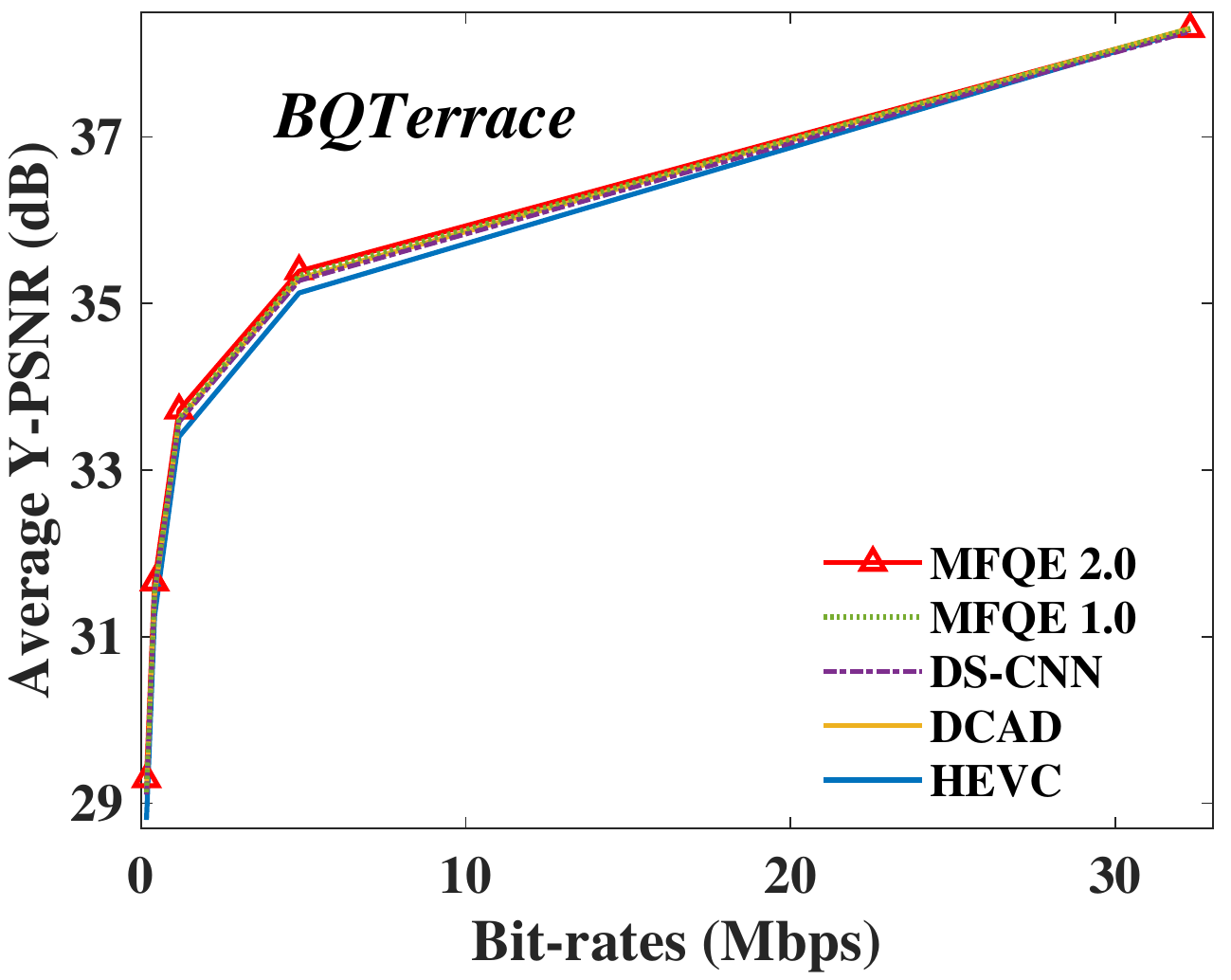}
	\end{minipage}
	\begin{minipage}[t]{1.73in}
		\centering
		\includegraphics[width=1.73in]{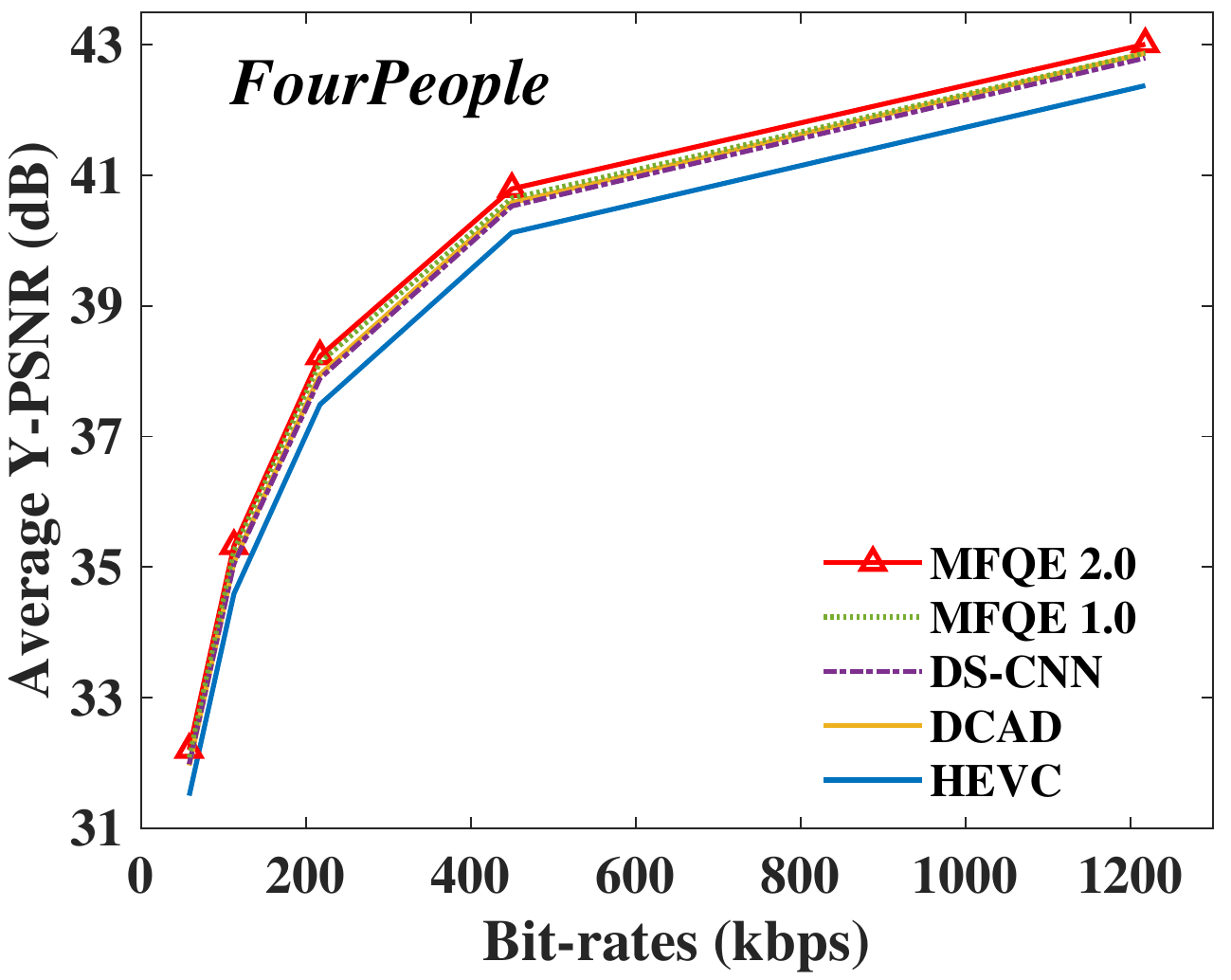}
	\end{minipage}
	
	\caption{Rate-distortion curves of four test sequences.}
	\label{bdpsnr}
	\vspace{-0.5em}
\end{figure}

\vspace{0.3cm}
\noindent\textbf{Overall quality enhancement. }		
Table\thinspace\ref{dpsnr} presents the results of $\Delta$PSNR and $\Delta$SSIM, averaged over all frames of each test sequence.
As shown in this table, our MFQE approach consistently outperforms all compared approaches.
To be specific, at QP = 37, the highest $\Delta$PSNR of our MFQE approach reaches 0.920 dB, i.e., for sequence \textit{PeopleOnStreet}.
The averaged $\Delta$PSNR of our MFQE approach is 0.562 dB, which is $23.5\%$ higher than that of MFQE 1.0 (0.455 dB), $88.0\%$ higher than that of Li \textit{et al.} (0.299 dB), $74.5\%$ higher than that of DCAD (0.322 dB), and $87.3\%$ higher than that of DS-CNN (0.300 dB).
Even higher $\Delta$PSNR improvement can be observed, when compared with AR-CNN and DnCNN.
At other QPs (= 22, 27, 32 and 42), our MFQE approach consistently outperforms other state-of-the-art video quality enhancement approaches.
Similar improvement can be found for SSIM in Table\thinspace\ref{dpsnr}. This demonstrates the robustness of our MFQE approach in enhancing video quality.
This is mainly attributed to the significant improvement on the quality of non-PQFs, which is the majority of compressed video frames.

\vspace{0.3cm}
\noindent\textbf{Rate-distortion performance.} We further evaluate the rate-distortion performance of our MFQE approach by comparing with other approaches.
First, Fig.\thinspace\ref{bdpsnr} shows the rate-distortion curves of our and other state-of-the-art approaches over four selected sequences.
Note that the results of the DCAD and DS-CNN approaches are plotted in this figure, since they perform better than other compared approaches.
We can see from Fig.\thinspace\ref{bdpsnr} that our MFQE approach performs better than other approaches in rate-distortion performance.
Then, we quantify the rate-distortion performance by evaluating the BD-bitrate (BD-BR) reduction, which is calculated over the PSNR results of five QPs (= 22, 27, 32, 37 and 42).
The results are presented in Table\thinspace\ref{tab_bdr}.
As can be seen, the BD-BR reduction of our MFQE approach is $14.06\%$ on average, while that of the second-best approach DCAD is only $8.89\%$ on average.
In general, the quality enhancement of our MFQE approach is equivalent to improving rate-distortion performance.

\begin{figure}[!t]
	\vspace{+0.0em}
	\centering
	\includegraphics[width = 1\linewidth]{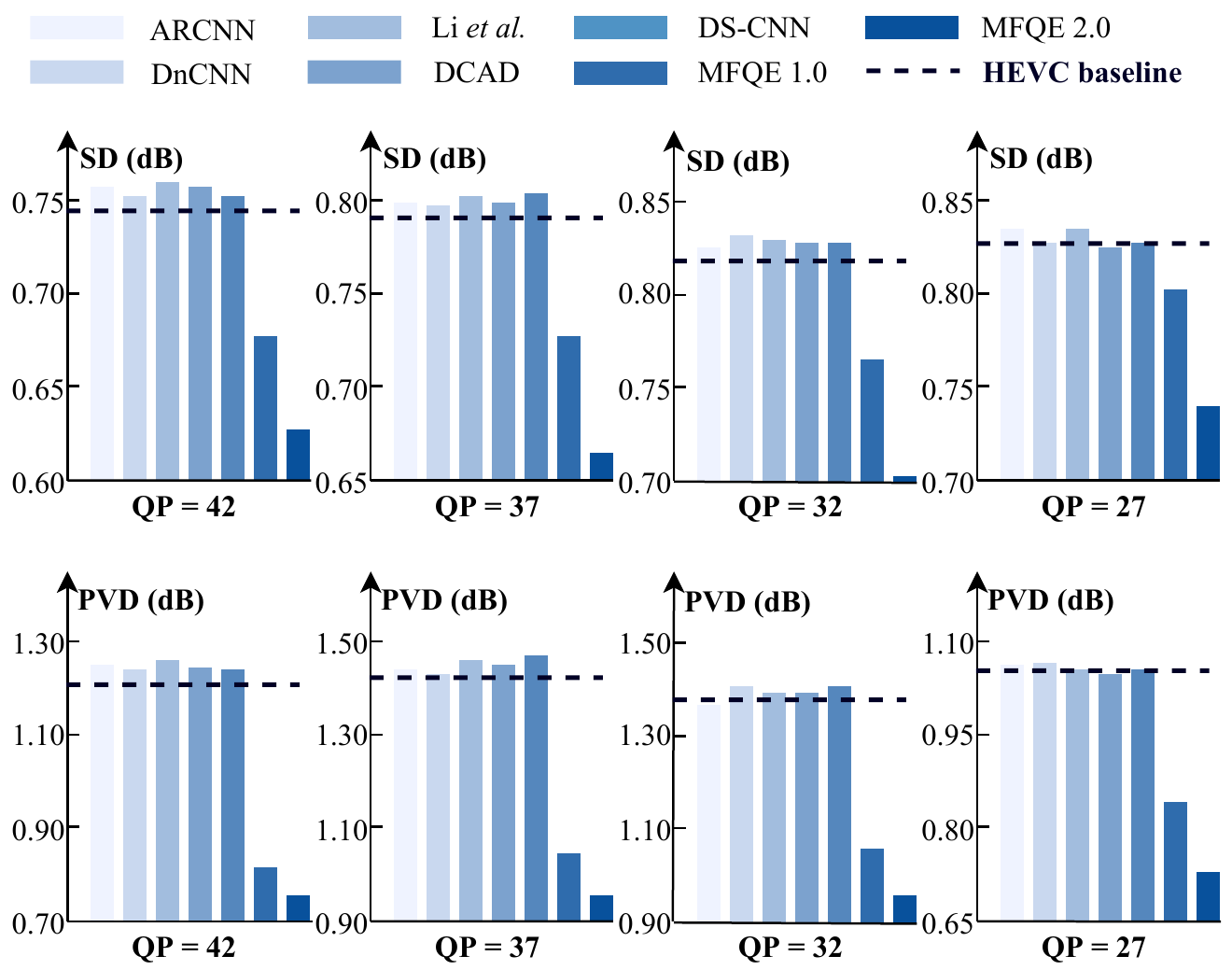}
	\caption{\footnotesize{Averaged SD and PVD of test sequences.}}
	\label{sdpvd}
	\vspace{-0.8em}
\end{figure}

\begin{figure}[!t]
	\vspace{+.0em}
	\centering
	\subfigure{
		\includegraphics[width=1\linewidth]{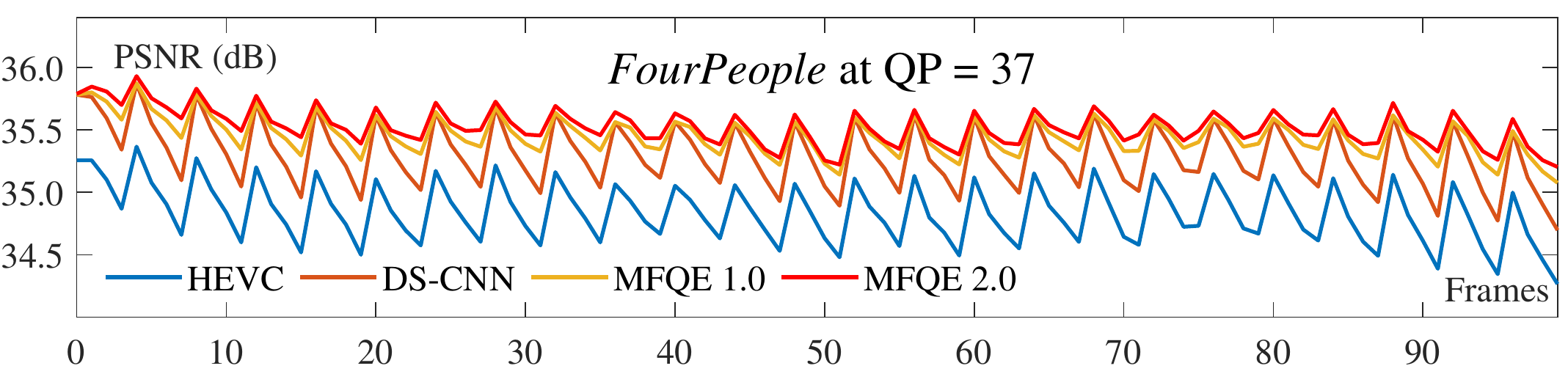}
	}
	\subfigure{
		\includegraphics[width=1\linewidth]{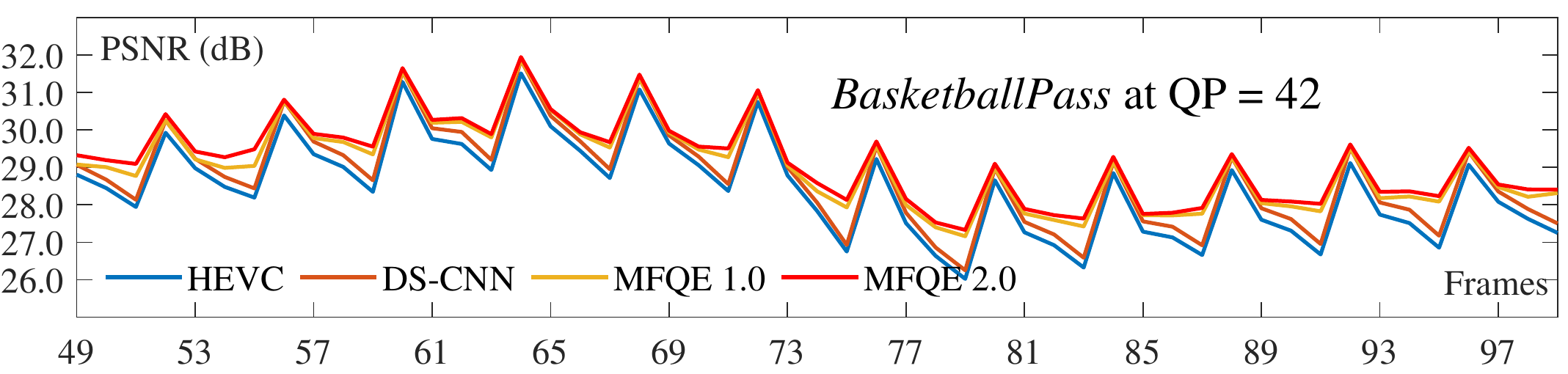}
	}
	\caption{\footnotesize{PSNR curves of HEVC baseline and our MFQE approach.}}
	\label{psnrcurveaf}
	\vspace{-0.5em}
\end{figure}

\vspace{0.3cm}
\noindent\textbf{Quality fluctuation. }		
Apart from the compression artifacts, the quality fluctuation in compressed videos may also lead to degradation of QoE \cite{He2001Low,Vito2005PSNR,Hu2012Adaptive}.
Fortunately, our MFQE approach is beneficial to mitigate the quality fluctuation, because of its significant quality improvement on non-PQFs as found in Fig.\thinspace\ref{deltapsnr}.
We evaluate the fluctuation of video quality in terms of the SD and PVD results of PSNR curves, which are introduced in Section\thinspace\ref{quality}.
Fig.\thinspace\ref{sdpvd} shows the SD and PVD values averaged over all 18 test sequences, which are obtained from the quality enhancement approaches and the HEVC baseline.
As shown in this figure, our MFQE approach succeeds in reducing the SD and PVD, while other five compared approaches enlarge the SD and PVD values over the HEVC baseline.
The reason is that our MFQE approach has considerably larger PSNR improvement for non-PQFs than that for PQFs, thus reducing the quality gap between PQFs and non-PQFs.
In addition, Fig.\thinspace\ref{psnrcurveaf} shows the PSNR curves of two selected test sequences, for our MFQE approach and the HEVC baseline.
It can be seen that the PSNR fluctuation of our MFQE approach is significantly smaller than the HEVC baseline.
In summary, our approach is also capable of reducing the quality fluctuation of video compression.

\begin{figure*}[!t]
	\vspace{-0.5em}
	\centering
	\includegraphics[width = 1\linewidth]{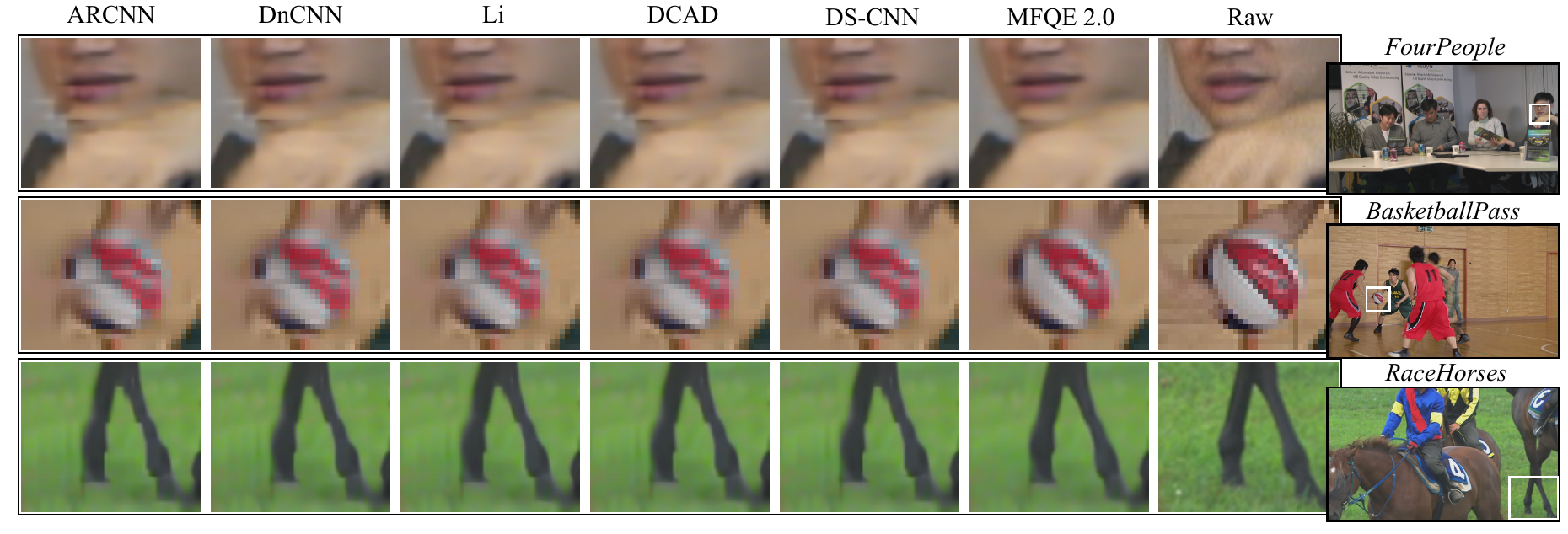}
	\caption{
		Subjective quality performance on \textit{Fourpeople} at QP = 37, \textit{BasketballPass} at QP = 37 and \textit{RaceHorses} at QP = 42.
	}
	\label{demo}
	\vspace{-0.0em}
\end{figure*}

\vspace{0.3cm}
\noindent\textbf{Subjective quality performance. }																																																															
Fig.\thinspace\ref{demo} shows the subjective quality performance on the sequences \textit{Fourpeople} at QP = 37, \textit{BasketballPass} at QP = 37 and \textit{RaceHorses} at QP = 42.
It can be observed that our MFQE approach reduces the compression artifacts much more effectively than other five compared approaches.
Specifically, the severely distorted content, e.g., the cheek in \textit{Fourpeople}, the ball in \textit{BasketballPass} and the horse's feet in \textit{RaceHorses}, can be finely restored by our MFQE approach with multi-frame strategy.
By contrast, such compression distortion can hardly be restored by the compared approaches, as they only use the single low-quality frame.
Therefore, our MFQE approach also performs well in subjective quality enhancement.

\begin{table}[!t]
	\vspace{+0.0em}
	\renewcommand\arraystretch{1.1}
	\centering
	\scriptsize
	\caption{Test speed (fps) and parameters.}
	
	\begin{threeparttable}
		\begin{tabular}{l l p{0.7cm}<{\centering} p{0.7cm}<{\centering} p{0.4cm}<{\centering} p{0.4cm}<{\centering} p{0.4cm}<{\centering} r}
			
			\toprule
			\multicolumn{2}{l}{\multirow{2}{*}{\textbf{MFQE}}} &  \multicolumn{5}{c}{\textbf{Test speed}} & \multirow{2}{*}{\textbf{Parameters}} \\
			
			\cline{3-7}
			&  & WQXGA & 1080p & 480p & 240p & 720p & \\
			
			\midrule		
			\multirow{2}{*}{1.0} & DS-CNN\tnote{1} & 0.57 & 1.12 & 5.92 & 19.38 & 2.54 & 1,344,449  \\[-0.1em]
			
			& MF-CNN\tnote{2} & 0.36 & 0.73 & 3.83 & 12.55 & 1.63 & 1,787,547 \\
			
			\midrule			
			
			2.0 & MF-CNN\tnote{3} & \textbf{0.79} & \textbf{1.61} & \textbf{8.35} & \textbf{25.29} & \textbf{3.66} & \textbf{255,422} \\
			
			\bottomrule
			
		\end{tabular}
		
		\begin{tablenotes}
			\footnotesize
			\item[1] for PQF enhancement.
			\item[2] for non-PQF enhancement.
			\item[3] for both PQF and non-PQF enhancement.
		\end{tablenotes}
		
	\end{threeparttable}
	
	\label{tab_complex}
	\vspace{-0.0em}
\end{table}

\vspace{0.3cm}
\noindent\textbf{Test speed. }		
We evaluate the test speed of quality enhancement using a computer equipped with a CPU of Intel i7-8700 3.20GHz and a GPU of GeForce GTX 1080 Ti.
Specifically, we measure the average frame per second (fps), when testing video sequences at different resolutions.
Note that the test set has been divided into 5 classes at different resolutions in \cite{ohm2012comparison}.
The results averaged over sequences at different resolutions are reported in Table\thinspace\ref{tab_complex}.
As shown in this table, when enhancing non-PQFs, MFQE 2.0 can achieve at least 2 times acceleration compared to MFQE 1.0.
For PQFs, MFQE 2.0 is also considerably faster than MFQE 1.0.
The reason is that the parameters of the MF-CNN architecture in MFQE 2.0 are significantly fewer than those in MFQE 1.0.
In a word, MFQE 2.0 is efficient in video quality enhancement, and its efficiency is mainly due to its lightweight structure.

Furthermore, we calculate the number of operations for the MFQE approach.
For MFQE 1.0, there are 99,561 additions and 215,150,624 multiplications needed for enhancing a 64$\times$64 patch, while those for MFQE 2.0 are 150,276 and 5,942,640.
The reason for the dramatic reduction of operations is that we decrease the number of filters in the mapping structure of MF-CNN from 64 to 32, and relieve the burden of feature extraction by cutting the number of output feature maps from 128 to 32.
At the same time, we deepen the mapping structure and introduce the dense strategy, batch normalization and residual learning.
This way, the nonlinearity of MF-CNN is largely improved, while the number of parameters is effectively saved.
In a word, MFQE 2.0 is efficient in video quality enhancement, and its efficiency is mainly due to the lightweight structure.

\subsection{Ablation study}
\label{Ablate}

\begin{figure}[!t]
	\vspace{+0.0em}
	\centering
	\includegraphics[width = 3.6in]{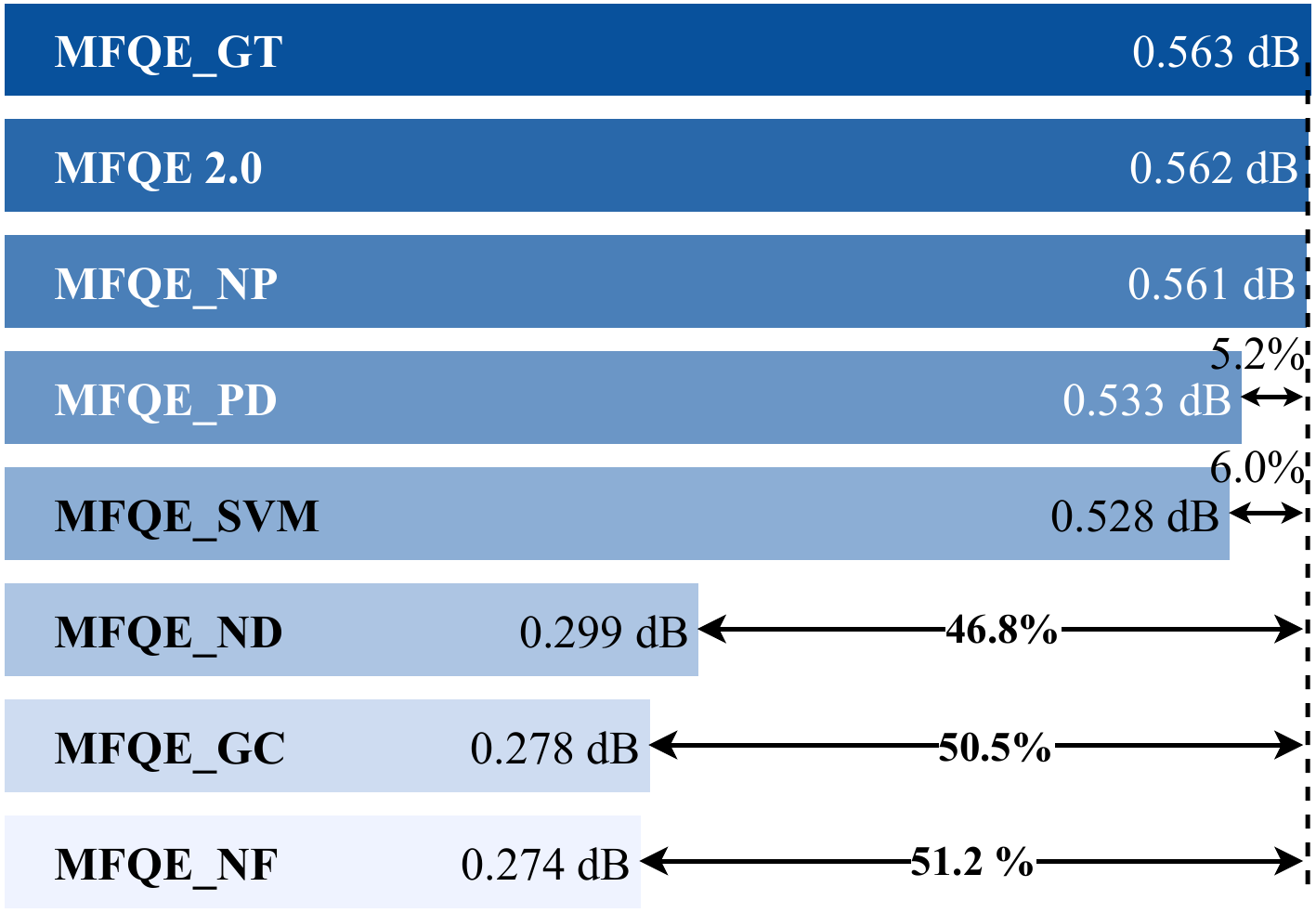}
	\caption{Overall $\Delta$PSNR (dB) of test sequences in ablation study. The explanations of abbreviations are as follows:
		(1) GT: Ground Truth PQFs. (2) NP: No Post-processing. (3) PD: Previous Database. (4) SVM: SVM-based detector. (5) ND: No Dense connection. (6) GC: General CNN. (7) NF: neighboring Frames serving as ``PQFs''.}
	\label{fig_ablate}
	\vspace{-0.0em}
\end{figure}

\noindent\textbf{PQF detector. }
In this section, we validate the necessity and effectiveness of utilizing PQFs to enhance the quality of non-PQFs.
To this end, we retrain the MF-CNN model of our MFQE approach to enhance non-PQFs with the help of adjacent frames, instead of PQFs.
The MF-CNN network and experiment settings are all consistent with those in Sections \ref{mc} and \ref{setting}.
The retrained model is represented by MFQE\_NF (i.e., MFQE with neighboring Frames), and the experimental results are shown in Fig.\thinspace\ref{fig_ablate}, which are obtained by averaging over all 18 test sequences compressed at QP = 37.
We can see that our approach without considering PQFs can only result in 0.274 dB for $\Delta$PSNR gain.
By contrast, as aforementioned,  our approach with PQFs can achieve 0.562 dB enhancement in $\Delta$PSNR.
Moreover, as validated in Section\thinspace\ref{MFQEperformance}, our MFQE approach obtains considerably higher enhancement on non-PQFs, when compared to the single-frame approaches.
In a word, the above ablation study demonstrates the  necessity and effectiveness of utilizing PQFs in the video quality enhancement task.

Besides, we test the MF-CNN model with ground truth PQFs.
Specifically, the ground truth PQF labels are obtained according to the PSNR curves and the definition of PQFs.
The experimental results (denoted by MFQE\_GT, i.e., MFQE with Ground Truth PQFs) are shown in Fig.\thinspace\ref{fig_ablate}.
As we can see, the average $\Delta$PSNR is 0.563 dB.
This indicates an upper bound on the performance with respect to PQF estimation.

Also, we test the impact of post-processing of the PQF detector, i.e., removing the neighboring PQFs and inserting PQFs between two PQFs with long distance.
Specifically, we test the $F_1$-score of the PQF detector without post-processing, and further evaluate its performance on quality enhancement (denoted by MFQE\_NP, i.e., MFQE with No Post-processing) in terms of $\Delta$PSNR.
The average $F_1$-score with post-processing slightly increases from 98.15$\%$ to 98.21$\%$ compared to the detector without post-processing.
Additionally, the average $\Delta$PSNR decreases by 0.001 dB after removing post-processing.
Although the $\Delta$PSNR improvement by taking post-processing is minor, the post-processing is still necessary in some extreme cases, where post-processing can prevent MFQE approach from inaccuate motion compensation and inferior quality enhancement.
Take sequence \textit{KristenAndSara} as an example.
The non-PQF labels of frames 273 and 277 are corrected to PQFs.
Consequently, the average $\Delta$PSNR of frames 270 to 280 can increase from 0.659 dB to 0.724 dB after using post-processed labels.

Finally, we conduct an experiment to validate the improvement of quality enhancement after replacing SVM with BiLSTM both in training and evaluation, which is an advancement of MFQE 2.0 over MFQE 1.0.
Specifically, we first replace the BiLSTM detector of MFQE 2.0 with SVM in the detection stage.
Then, we retrain and test the model (denoted by MFQE\_SVM) which consists of the SVM based detector and MF-CNN.
The average $\Delta$PSNR decreases from 0.562 dB to 0.528 dB (i.e., 6.0\% degradation).
This validates the contribution of the improved PQF detector.

\noindent\textbf{Multi-scale and dense connection strategy. }
We further validate the effectiveness of the multi-scale feature extraction strategy and the densely connected structure in enhancing video quality.
First, we ablate all dense connections in the QE-subnet of our MFQE approach.
In addition, we increase the filter number of C11 from 32 to 50, so that the number of trainable parameters can be maintained for fair comparison.
The corresponding retrained model is denoted by MFQE\_ND (i.e., MFQE with No Dense connection).
Second, we ablate the multi-scale structure in the QE-subnet.
Based on the dense-ablated network above, we fix all kernel sizes of the feature extraction component to $5 \times 5$.
Other parts of the MFQE approach and experiment settings are all the same as those in Sections \ref{MFQE} and \ref{setting}.
Accordingly, the retrained model is represented as MFQE\_GC (i.e., MFQE with General CNN).
Fig.\thinspace\ref{fig_ablate} shows the ablation results, which are also averaged over all 18 test sequences at QP = 37.
As seen in this table, the PSNR improvement decreases from 0.562 dB to 0.299 dB (i.e., $46.8\%$ degradation) when disabling the dense connections, and then it reduces to 0.278 dB (i.e, $50.5\%$ degradation) when further ablating the multi-scale structure.
This indicates the effectiveness of our multi-scale strategy and the densely connected structure.

\vspace{0.3cm}
\noindent\textbf{Enlarged database. }
One of the contributions in this paper is that we enlarge our database from 70 to 160 uncompressed video sequences.
Here, we verify the effectiveness of the enlarged database over our previous database \cite{yang2018multi}.
Specifically, we test the performance of our MFQE approach trained over the database in \cite{yang2018multi}.
Then, the performance is evaluated on all 18 test sequences at QP = 37.
The retrained model with its corresponding test result is represented by MFQE\_PD (i.e., MFQE with the Previous Database) in Fig.\thinspace\ref{fig_ablate}.
We can see that MFQE 2.0 achieves substantial improvement on quality enhancement compared with MFQE-A7.
In particular, the performance of MFQE 2.0 improves $\Delta$PSNR from 0.533 dB to 0.562 dB on average.
Hence, our enlarged database is effective in improving video quality enhancement performance.

\subsection{Generalization ability of our MFQE approach}
\label{trans}
\vspace{0.3cm}
\noindent\textbf{Transfer to H.264. }
We verify the generalization ability of our MFQE approach for video sequences compressed by another standard.
To this end, we test our MFQE approach on the 18 test sequences compressed by H.264 at QP = 37.
Note that the test model is the same as that in Section 5.3, which is trained over the training set compressed by HEVC at QP = 37.
Consequently, the average PSNR improvement is 0.422 dB.
Also, we test the performance of MFQE model retrained over H.264 dataset.
The average PSNR improvement is 0.464 dB.
In a word, the MFQE model trained over HEVC dataset performs well on H.264 videos, and the MFQE model retrained on H.264 can slightly improve the performance of quality enhancement.
This implies the high generalization ability of our MFQE approach across different compression standards.

\vspace{0.3cm}
\noindent\textbf{Performance on other sequences. }
It is worth mentioning that the test set in \cite{yang2018multi} is different from that in this paper.
In our previous work \cite{yang2018multi}, 10 test sequences are randomly selected from the previous database including 70 videos.
In this paper, our 18 test sequences are selected by Joint Collaborative Team on Video Coding (JCT-VC)\cite{ohm2012comparison}, which is a standard test set for video compression.
For fair comparison, we test the performance of our MFQE 2.0 and all compared approaches over the previous test set. The experimental results are presented in Table\thinspace\ref{conf10}.
Note that 4 test sequences among the 10 test sequences overlap with the 18 test sequences of the above experiments.
We can see from Table\thinspace\ref{conf10} that our approach has 0.680 dB improvement in $\Delta$PSNR and again outperforms other approaches.
In this table, the results of compared approaches are also better than those reported in \cite{yang2018multi} and their papers.
It is because of retraining over the enlarged database.
In conclusion, our MFQE approach has high generalization ability over different test sequences.

\begin{table}[!t]
	\vspace{+0.5em}
	\renewcommand\arraystretch{1.1}
	\centering
	\scriptsize
	\caption{Overall $\Delta$PSNR (dB) of 10 test sequences at QP = 37.}
	
	\begin{tabular}{p{0.3cm}<{\centering} p{0.65cm}<{\centering} p{0.65cm}<{\centering} p{0.65cm}<{\centering} p{0.65cm}<{\centering} p{0.65cm}<{\centering} p{0.65cm}<{\centering} p{0.65cm}<{\centering}}
		
		\toprule
		\multirow{2}{*}{Seq.} & AR-CNN & \multirow{2}{*}{DnCNN} & Li \textit{et al.} & \multirow{2}{*}{DCAD} & DS-CNN & MFQE 1.0 & MFQE 2.0 \\
		
		\midrule
		1 & 0.280 & 0.359 & 0.459 & 0.510 & 0.415 & 0.655 & \textbf{0.775}\\
		
		2 & 0.266 & 0.303 & 0.387 & 0.399 & 0.339 & 0.492 & \textbf{0.579}\\
		
		3 & 0.315 & 0.365 & 0.422 & 0.439 & 0.394 & 0.629 & \textbf{0.735}\\
		
		4 & 0.321 & 0.312 & 0.401 & 0.421 & 0.388 & 0.599 & \textbf{0.719}\\
		
		5 & 0.237 & 0.229 & 0.287 & 0.311 & 0.290 & 0.414 & \textbf{0.476}\\
		
		6 & 0.261 & 0.312 & 0.392 & 0.373 & 0.343 & 0.659 & \textbf{0.723}\\
		
		7 & 0.346 & 0.414 & 0.482 & 0.481 & 0.465 & 0.772 & \textbf{0.920}\\
		
		8 & 0.219 & 0.244 & 0.187 & 0.279 & 0.280 & 0.472 & \textbf{0.550}\\
		
		9 & 0.267 & 0.311 & 0.328 & 0.317 & 0.358 & 0.394 & \textbf{0.594}\\
		
		10 & 0.259 & 0.307 & 0.343 & 0.332 & 0.375 & 0.484 & \textbf{0.728}\\
		
		\midrule
		Ave. & 0.277 & 0.316 & 0.369 & 0.386 & 0.365 & 0.557 & \textbf{0.680}\\
		
		\midrule
		\multicolumn{8}{c}{1: \textit{TunnelFlag} 2: \textit{BarScene} 3: \textit{Vidyo1} 4: \textit{Vidyo3} 5: \textit{Vidyo4} 6: \textit{MaD}} \\
		
		\multicolumn{8}{c}{7: \textit{PeopleOnStreet} 8: \textit{Kimono} 9: \textit{RaceHorses} 10: \textit{BasketballPass}}
		
	\end{tabular}
	
	\label{conf10}
	\vspace{-1.0em}
\end{table}

\section{Conclusion}

In this paper, we have proposed a CNN-based MFQE approach to enhance the quality of compressed video by reducing compression artifacts.
Differing from the conventional single-frame quality enhancement approaches, our MFQE approach improves the quality of one frame by utilizing its nearest PQFs that have higher quality.
To this end, we developed a BiLSTM-based PQF detector to classify PQFs and non-PQFs in compressed video.
Then, we proposed a novel CNN framework, called MF-CNN, to enhance the quality of non-PQFs.
Specifically, our MF-CNN framework consists of two subnets, i.e., the MC-subnet and QE-subnet.
First, the MC-subnet compensates motion between PQFs and non-PQFs.
Subsequently, the QE-subnet enhances the quality of each non-PQF by feeding the current non-PQF and the nearest compensated PQFs.
In addition, PQF quality is enhanced in the same way.
Finally, extensive experimental results showed that our MFQE approach significantly improves the quality of compressed video, superior to other state-of-the-art approaches.
Consequently, the overall quality can be significantly enhanced, with considerably higher quality and less quality fluctuation than other approaches.

There may exist two research directions for future work. (1) Our work in this paper only takes PSNR and SSIM as the objective metrics to be enhanced. The potential future work may further embrace perceptual quality metrics in our approach to improve the Quality of Experience (QoE) in video quality enhancement. (2) Our work mainly focuses on the quality enhancement at the decoder side. To further improve the performance of quality enhancement, information from the encoder, such as the partition of coding units, can be utilized. This is a promising future work.

\section{Acknowledgment}

This work was supported by the NSFC projects 61876013, 61922009 and 61573037.

\ifCLASSOPTIONcaptionsoff
  \newpage
\fi

\bibliographystyle{IEEEtran}
\bibliography{egbib}

\begin{thebibliography}{10}
\providecommand{\url}[1]{#1}
\csname url@samestyle\endcsname
\providecommand{\newblock}{\relax}
\providecommand{\bibinfo}[2]{#2}
\providecommand{\BIBentrySTDinterwordspacing}{\spaceskip=0pt\relax}
\providecommand{\BIBentryALTinterwordstretchfactor}{4}
\providecommand{\BIBentryALTinterwordspacing}{\spaceskip=\fontdimen2\font plus
\BIBentryALTinterwordstretchfactor\fontdimen3\font minus
  \fontdimen4\font\relax}
\providecommand{\BIBforeignlanguage}[2]{{%
\expandafter\ifx\csname l@#1\endcsname\relax
\typeout{** WARNING: IEEEtran.bst: No hyphenation pattern has been}%
\typeout{** loaded for the language `#1'. Using the pattern for}%
\typeout{** the default language instead.}%
\else
\language=\csname l@#1\endcsname
\fi
#2}}
\providecommand{\BIBdecl}{\relax}
\BIBdecl

\bibitem{Cisco}
I.~Cisco~Systems, ``Cisco visual networking index: Global mobile data traffic
  forecast update,''
  https://www.cisco.com/c/en/us/solutions/collateral/service-provider/visual-networking-index-vni/mobile-white-paper-c11-520862.html.

\bibitem{seshadrinathan2010study}
K.~Seshadrinathan, R.~Soundararajan, A.~C. Bovik, and L.~K. Cormack, ``Study of
  subjective and objective quality assessment of video,'' \emph{IEEE
  transactions on image processing}, vol.~19, no.~6, pp. 1427--1441, 2010.

\bibitem{li2015weight}
S.~Li, M.~Xu, X.~Deng, and Z.~Wang, ``Weight-based r-$\lambda$ rate control for
  perceptual hevc coding on conversational videos,'' \emph{Signal Processing:
  Image Communication}, vol.~38, pp. 127--140, 2015.

\bibitem{tan2016video}
T.~K. Tan, R.~Weerakkody, M.~Mrak, N.~Ramzan, V.~Baroncini, J.-R. Ohm, and
  G.~J. Sullivan, ``Video quality evaluation methodology and verification
  testing of hevc compression performance,'' \emph{IEEE Transactions on
  Circuits and Systems for Video Technology}, vol.~26, no.~1, pp. 76--90, 2016.

\bibitem{bampis2017study}
C.~G. Bampis, Z.~Li, A.~K. Moorthy, I.~Katsavounidis, A.~Aaron, and A.~C.
  Bovik, ``Study of temporal effects on subjective video quality of
  experience,'' \emph{IEEE Transactions on Image Processing}, vol.~26, no.~11,
  pp. 5217--5231, 2017.

\bibitem{yang2018saliency}
R.~Yang, M.~Xu, Z.~Wang, Y.~Duan, and X.~Tao, ``Saliency-guided complexity
  control for hevc decoding,'' \emph{IEEE Transactions on Broadcasting}, 2018.

\bibitem{gupta2005restoration}
M.~D. Gupta, S.~Rajaram, N.~Petrovic, and T.~S. Huang, ``Restoration and
  recognition in a loop,'' in \emph{Computer Vision and Pattern Recognition,
  2005. CVPR 2005. IEEE Computer Society Conference on}, vol.~1.\hskip 1em plus
  0.5em minus 0.4em\relax IEEE, 2005, pp. 638--644.

\bibitem{hennings2008simultaneous}
P.~H. Hennings-Yeomans, S.~Baker, and B.~V. Kumar, ``Simultaneous
  super-resolution and feature extraction for recognition of low-resolution
  faces,'' in \emph{Computer Vision and Pattern Recognition, 2008. CVPR 2008.
  IEEE Conference on}.\hskip 1em plus 0.5em minus 0.4em\relax IEEE, 2008, pp.
  1--8.

\bibitem{nishiyama2009facial}
M.~Nishiyama, H.~Takeshima, J.~Shotton, T.~Kozakaya, and O.~Yamaguchi, ``Facial
  deblur inference to improve recognition of blurred faces,'' in \emph{Computer
  Vision and Pattern Recognition, 2009. CVPR 2009. IEEE Conference on}.\hskip
  1em plus 0.5em minus 0.4em\relax IEEE, 2009, pp. 1115--1122.

\bibitem{zhang2011close}
H.~Zhang, J.~Yang, Y.~Zhang, N.~M. Nasrabadi, and T.~S. Huang, ``Close the
  loop: Joint blind image restoration and recognition with sparse
  representation prior,'' in \emph{Computer Vision (ICCV), 2011 IEEE
  International Conference on}.\hskip 1em plus 0.5em minus 0.4em\relax IEEE,
  2011, pp. 770--777.

\bibitem{liew2004blocking}
A.-C. Liew and H.~Yan, ``Blocking artifacts suppression in block-coded images
  using overcomplete wavelet representation,'' \emph{IEEE Transactions on
  Circuits and Systems for Video Technology}, vol.~14, no.~4, pp. 450--461,
  2004.

\bibitem{foi2007pointwise}
A.~Foi, V.~Katkovnik, and K.~Egiazarian, ``Pointwise shape-adaptive {DCT} for
  high-quality denoising and deblocking of grayscale and color images,''
  \emph{IEEE Transactions on Image Processing}, vol.~16, no.~5, pp. 1395--1411,
  2007.

\bibitem{wang2013adaptive}
C.~Wang, J.~Zhou, and S.~Liu, ``Adaptive non-local means filter for image
  deblocking,'' \emph{Signal Processing: Image Communication}, vol.~28, no.~5,
  pp. 522--530, 2013.

\bibitem{jancsary2012loss}
J.~Jancsary, S.~Nowozin, and C.~Rother, ``Loss-specific training of
  non-parametric image restoration models: A new state of the art,'' in
  \emph{Proceedings of the European Conference on Computer Vision
  (ECCV)}.\hskip 1em plus 0.5em minus 0.4em\relax Springer, 2012, pp. 112--125.

\bibitem{jung2012image}
C.~Jung, L.~Jiao, H.~Qi, and T.~Sun, ``Image deblocking via sparse
  representation,'' \emph{Image Communication}, vol.~27, no.~6, pp. 663--677,
  2012.

\bibitem{chang2014reducing}
H.~Chang, M.~K. Ng, and T.~Zeng, ``Reducing artifacts in {JPEG} decompression
  via a learned dictionary,'' \emph{IEEE Transactions on Signal Processing},
  vol.~62, no.~3, pp. 718--728, 2014.

\bibitem{dong2015compression}
C.~Dong, Y.~Deng, C.~Change~Loy, and X.~Tang, ``Compression artifacts reduction
  by a deep convolutional network,'' in \emph{Proceedings of the IEEE
  International Conference on Computer Vision (ICCV)}, 2015, pp. 576--584.

\bibitem{Guo2016Building}
J.~Guo and H.~Chao, ``Building dual-domain representations for compression
  artifacts reduction,'' in \emph{Proceedings of the European Conference on
  Computer Vision (ECCV)}, 2016, pp. 628--644.

\bibitem{wang2016d3}
Z.~Wang, D.~Liu, S.~Chang, Q.~Ling, Y.~Yang, and T.~S. Huang, ``{D}3: Deep
  dual-domain based fast restoration of {JPEG}-compressed images,'' in
  \emph{Proceedings of the IEEE Conference on Computer Vision and Pattern
  Recognition (CVPR)}, 2016, pp. 2764--2772.

\bibitem{Zhang2017Beyond}
K.~Zhang, W.~Zuo, Y.~Chen, D.~Meng, and L.~Zhang, ``Beyond a gaussian denoiser:
  Residual learning of deep cnn for image denoising,'' \emph{IEEE Transactions
  on Image Processing}, vol.~26, no.~7, pp. 3142--3155, 2017.

\bibitem{li2017efficient}
K.~Li, B.~Bare, and B.~Yan, ``An efficient deep convolutional neural networks
  model for compressed image deblocking,'' in \emph{Proceedings of the IEEE
  International Conference on Multimedia and Expo (ICME)}.\hskip 1em plus 0.5em
  minus 0.4em\relax IEEE, 2017, pp. 1320--1325.

\bibitem{Cavigelli2017CAS}
L.~Cavigelli, P.~Hager, and L.~Benini, ``{CAS-CNN}: A deep convolutional neural
  network for image compression artifact suppression,'' in \emph{Proceedings of
  the International Joint Conference on Neural Networks (IJCNN)}, 2017, pp.
  752--759.

\bibitem{tai2017memnet}
Y.~Tai, J.~Yang, X.~Liu, and C.~Xu, ``Memnet: A persistent memory network for
  image restoration,'' in \emph{Proceedings of the IEEE Conference on Computer
  Vision and Pattern Recognition (CVPR)}, 2017, pp. 4539--4547.

\bibitem{yang2017decoder}
R.~Yang, M.~Xu, and Z.~Wang, ``Decoder-side {HEVC} quality enhancement with
  scalable convolutional neural network,'' in \emph{Multimedia and Expo (ICME),
  2017 IEEE International Conference on}.\hskip 1em plus 0.5em minus
  0.4em\relax IEEE, 2017, pp. 817--822.

\bibitem{yang2018enhancing}
R.~Yang, M.~Xu, T.~Liu, Z.~Wang, and Z.~Guan, ``Enhancing quality for hevc
  compressed videos,'' \emph{IEEE Transactions on Circuits and Systems for
  Video Technology}, 2018.

\bibitem{lecun1998gradient}
Y.~LeCun, L.~Bottou, Y.~Bengio, and P.~Haffner, ``Gradient-based learning
  applied to document recognition,'' \emph{Proceedings of the IEEE}, vol.~86,
  no.~11, pp. 2278--2324, 1998.

\bibitem{Kappeler2016Video}
A.~Kappeler, S.~Yoo, Q.~Dai, and A.~K. Katsaggelos, ``Video super-resolution
  with convolutional neural networks,'' \emph{IEEE Transactions on
  Computational Imaging}, vol.~2, no.~2, pp. 109--122, 2016.

\bibitem{Li2017Video}
D.~Li and Z.~Wang, ``Video super-resolution via motion compensation and deep
  residual learning,'' \emph{IEEE Transactions on Computational Imaging},
  vol.~PP, no.~99, pp. 1--1, 2017.

\bibitem{Caballero_2017_CVPR}
J.~Caballero, C.~Ledig, A.~Aitken, A.~Acosta, J.~Totz, Z.~Wang, and W.~Shi,
  ``Real-time video super-resolution with spatio-temporal networks and motion
  compensation,'' in \emph{Proceedings of the IEEE Conference on Computer
  Vision and Pattern Recognition (CVPR)}, July 2017.

\bibitem{yang2018multi}
R.~Yang, M.~Xu, Z.~Wang, and T.~Li, ``Multi-frame quality enhancement for
  compressed video,'' in \emph{Proceedings of the IEEE Conference on Computer
  Vision and Pattern Recognition (CVPR)}, 2018, pp. 6664--6673.

\bibitem{Ioffe2015Batch}
S.~Ioffe and C.~Szegedy, ``Batch normalization: Accelerating deep network
  training by reducing internal covariate shift,'' \emph{arXiv preprint
  arXiv:1502.03167}, 2015.

\bibitem{huang2017densely}
G.~Huang, Z.~Liu, L.~Van Der~Maaten, and K.~Q. Weinberger, ``Densely connected
  convolutional networks.'' in \emph{CVPR}, vol.~1, no.~2, 2017, p.~3.

\bibitem{ohm2012comparison}
J.-R. Ohm, G.~J. Sullivan, H.~Schwarz, T.~K. Tan, and T.~Wiegand, ``Comparison
  of the coding efficiency of video coding standards—including high
  efficiency video coding (hevc),'' \emph{IEEE Transactions on circuits and
  systems for video technology}, vol.~22, no.~12, pp. 1669--1684, 2012.

\bibitem{dai2017convolutional}
Y.~Dai, D.~Liu, and F.~Wu, ``A convolutional neural network approach for
  post-processing in {hevc} intra coding,'' in \emph{Proceedings of the
  International Conference on Multimedia Modeling (MMM)}.\hskip 1em plus 0.5em
  minus 0.4em\relax Springer, 2017, pp. 28--39.

\bibitem{Wang2017A}
T.~Wang, M.~Chen, and H.~Chao, ``A novel deep learning-based method of
  improving coding efficiency from the decoder-end for {HEVC},'' in
  \emph{Proceedings of the Data Compression Conference (DCC)}, 2017.

\bibitem{brandi2008super}
F.~Brandi, R.~de~Queiroz, and D.~Mukherjee, ``Super resolution of video using
  key frames,'' in \emph{Proceedings of the IEEE International Symposium on
  Circuits and Systems (ISCAS)}.\hskip 1em plus 0.5em minus 0.4em\relax IEEE,
  2008, pp. 1608--1611.

\bibitem{song2011video}
B.~C. Song, S.-C. Jeong, and Y.~Choi, ``Video super-resolution algorithm using
  bi-directional overlapped block motion compensation and on-the-fly dictionary
  training,'' \emph{IEEE Transactions on Circuits and Systems for Video
  Technology}, vol.~21, no.~3, pp. 274--285, 2011.

\bibitem{huang2018video}
Y.~Huang, W.~Wang, and L.~Wang, ``Video super-resolution via bidirectional
  recurrent convolutional networks,'' \emph{IEEE transactions on pattern
  analysis and machine intelligence}, vol.~40, no.~4, pp. 1015--1028, 2018.

\bibitem{dong2016image}
C.~Dong, C.~C. Loy, K.~He, and X.~Tang, ``Image super-resolution using deep
  convolutional networks,'' \emph{IEEE transactions on pattern analysis and
  machine intelligence}, vol.~38, no.~2, pp. 295--307, 2016.

\bibitem{Dosovitskiy2015FlowNet}
A.~Dosovitskiy, P.~Fischery, E.~Ilg, P.~Hausser, C.~Hazirbas, V.~Golkov,
  P.~V.~D. Smagt, D.~Cremers, and T.~Brox, ``Flow{N}et: Learning optical flow
  with convolutional networks,'' in \emph{Proceedings of the IEEE International
  Conference on Computer Vision (ICCV)}, 2015, pp. 2758--2766.

\bibitem{Ilg_2017_CVPR}
E.~Ilg, N.~Mayer, T.~Saikia, M.~Keuper, A.~Dosovitskiy, and T.~Brox,
  ``Flow{N}et 2.0: Evolution of optical flow estimation with deep networks,''
  in \emph{Proceedings of the IEEE Conference on Computer Vision and Pattern
  Recognition (CVPR)}, July 2017.

\bibitem{Makansi2017End}
O.~Makansi, E.~Ilg, and T.~Brox, ``End-to-end learning of video
  super-resolution with motion compensation,'' in \emph{Proceedings of the
  German Conference on Pattern Recognition (GCPR)}, 2017, pp. 203--214.

\bibitem{Shi2016Real}
W.~Shi, J.~Caballero, F.~Huszar, J.~Totz, A.~P. Aitken, R.~Bishop, D.~Rueckert,
  and Z.~Wang, ``Real-time single image and video super-resolution using an
  efficient sub-pixel convolutional neural network,'' in \emph{Proceedings of
  the IEEE Conference on Computer Vision and Pattern Recognition (CVPR)}, 2016,
  pp. 1874--1883.

\bibitem{Tao2017Detail}
X.~Tao, H.~Gao, R.~Liao, J.~Wang, and J.~Jia, ``Detail-revealing deep video
  super-resolution,'' in \emph{Proceedings of the IEEE Conference on Computer
  Vision and Pattern Recognition}, 2017, pp. 4472--4480.

\bibitem{Xiph}
Xiph.org, ``Xiph.org video test media,'' https://media.xiph.org/video/derf/.

\bibitem{VQEG}
VQEG, ``{VQEG} video datasets and organizations,''
  https://www.its.bldrdoc.gov/vqeg/video-datasets-and-organizations.aspx.

\bibitem{bossen2011common}
F.~Bossen, ``Common test conditions and software reference configurations,'' in
  \emph{Joint Collaborative Team on Video Coding (JCT-VC) of ITU-T SG16 WP3 and
  ISO/IEC JTC1/SC29/WG11, 5th meeting, Jan. 2011}, 2011.

\bibitem{le1992mpeg}
D.~J. Le~Gall, ``The mpeg video compression algorithm,'' \emph{Signal
  Processing: Image Communication}, vol.~4, no.~2, pp. 129--140, 1992.

\bibitem{Schafer1995Digital}
R.~Schafer and T.~Sikora, ``Digital video coding standards and their role in
  video communications,'' \emph{Proceedings of the IEEE}, vol.~83, no.~6, pp.
  907--924, 1995.

\bibitem{Sikora2002The}
T.~Sikora, ``The {MPEG}-4 video standard verification model,'' \emph{IEEE
  Transactions on Circuits and Systems for Video Technology}, vol.~7, no.~1,
  pp. 19--31, 2002.

\bibitem{wiegand2003overview}
T.~Wiegand, G.~J. Sullivan, G.~Bjontegaard, and A.~Luthra, ``Overview of the
  {H. 264/AVC} video coding standard,'' \emph{IEEE Transactions on Circuits and
  Systems for Video Technology}, vol.~13, no.~7, pp. 560--576, 2003.

\bibitem{sullivan2012overview}
G.~J. Sullivan, J.~Ohm, W.-J. Han, and T.~Wiegand, ``Overview of the high
  efficiency video coding ({HEVC}) standard,'' \emph{IEEE Transactions on
  Circuits and Systems for Video Technology}, vol.~22, no.~12, pp. 1649--1668,
  2012.

\bibitem{Hochreiter2001A}
S.~Hochreiter and M.~C. Mozer, \emph{A Discrete Probabilistic Memory Model for
  Discovering Dependencies in Time}.\hskip 1em plus 0.5em minus 0.4em\relax
  Springer Berlin Heidelberg, 2001.

\bibitem{Kingma2014Adam}
D.~P. Kingma and J.~Ba, ``Adam: A method for stochastic optimization,''
  \emph{Computer Science}, 2014.

\bibitem{He2001Low}
Z.~He, Y.~K. Kim, and S.~K. Mitra, ``Low-delay rate control for {DCT} video
  coding via $\rho$-domain source modeling,'' \emph{IEEE Transactions on
  Circuits and Systems for Video Technology}, vol.~11, no.~8, pp. 928--940,
  2001.

\bibitem{Vito2005PSNR}
F.~D. Vito and J.~C.~D. Martin, ``Psnr control for {GOP}-level constant quality
  in {H}.264 video coding,'' in \emph{Proceedings of the IEEE International
  Symposium on Signal Processing and Information Technology}, 2005, pp.
  612--617.

\bibitem{Hu2012Adaptive}
S.~Hu, H.~Wang, and S.~Kwong, ``Adaptive quantization-parameter clip scheme for
  smooth quality in {H}.264/{AVC},'' \emph{IEEE Transactions on Image
  Processing}, vol.~21, no.~4, pp. 1911--1919, 2012.

\end{thebibliography}

\begin{IEEEbiographynophoto}{Qunliang Xing}
	received the B.S. degree from Shenyuan Honors College and the School of Electronic and Information Engineering, Beihang University in 2019.
	He is currently pursuing the Ph.D. degree at the same university.
	His research interests mainly include image/video quality enhancement, video coding and computer vision.
\end{IEEEbiographynophoto}

\begin{IEEEbiographynophoto}{Zhenyu Guan}
	received the Ph.D. degree in Electronic Engineering from Imperial College London, UK in 2013.
	Since then, he has joined Beihang University as a Lecturer.
	He is a member of IEEE and IEICE. His current research interests include image processing and high performance computing.
	He has published more than 10 technical papers in international journals and conference proceedings, e.g., IEEE TIP.
\end{IEEEbiographynophoto}

\begin{IEEEbiographynophoto}{Mai Xu}
	(M'10, SM'16) received B.S. degree from Beihang University in 2003, M.S. degree from Tsinghua University in 2006 and Ph.D. degree from Imperial College London in 2010.
	From 2010-2012, he was working as a research fellow at Electrical Engineering Department, Tsinghua University.
	Since Jan. 2013, he has been with Beihang University as an Associate Professor.
	During 2014 to 2015, he was a visiting researcher of MSRA.
	His research interests mainly include image processing and computer vision.
	He has published more than 80 technical papers in international journals and conference proceedings, e.g., IEEE TPAMI, TIP, CVPR, ICCV and ECCV.
	He is the recipient of best paper awards of two IEEE conferences.
\end{IEEEbiographynophoto}

\begin{IEEEbiographynophoto}{Ren Yang}
	received the B.S. degree and the M.S. degree from the School of Electronic and  Information  Engineering,  Beihang University in 2016 and 2019, respectively.
	His research interests mainly include computer vision and video coding.
	He has published several papers in international journals and conference proceedings, e.g., IEEE TCSVT and CVPR.
\end{IEEEbiographynophoto}

\begin{IEEEbiographynophoto}{Tie Liu}
	is currently pursuing the B.S. degree at the School of Electronic and Information Engineering, Beihang University, Beijing, China. His research interests mainly include computer vision and video coding.
\end{IEEEbiographynophoto}

\begin{IEEEbiographynophoto}{Zulin Wang}
	(M'14) received the B.S. and M.S. degrees in electronic engineering from Beihang University, in 1986 and 1989, respectively.
	He received his Ph.D. degree at the same university in 2000.
	His research interests include image processing, electromagnetic countermeasure, and satellite communication technology.
	He is author or co-author of over 100 papers (including IEEE TPAMI, TIP and CVPR) and holds 6 patents, as well as published 2 books in these fields.
	He has undertaken approximately 30 projects related to image/video coding, image processing, etc.
\end{IEEEbiographynophoto}

\end{document}